\documentclass[acmsmall]{acmart}

\usepackage[normalem]{ulem}
\usepackage{latexsym}
\usepackage{microtype}

\usepackage{amsmath}
\usepackage{graphicx}

\usepackage{subfigure}
\usepackage{stfloats}

\usepackage{makecell}
\usepackage{multicol}
\usepackage{multirow}

\usepackage{tabularx}
\usepackage{color}

\usepackage{listings}
\AtBeginDocument{%
  \providecommand\BibTeX{{%
    \normalfont B\kern-0.5em{\scshape i\kern-0.25em b}\kern-0.8em\TeX}}}

\begin{document}
\settopmatter{printacmref=false} 
\renewcommand\footnotetextcopyrightpermission[1]{} 
\pagestyle{plain} 

\title{Building A Coding Assistant via the Retrieval-Augmented Language Model}

\newcommand\blfootnote[1]{%
\begingroup
\renewcommand\thefootnote{}\footnote{#1}%
\addtocounter{footnote}{-1}%
\endgroup
}
\blfootnote{This article is an extension of reference~\cite{li2023structure} on ACL 2023. The previous conference version focused only on learning the representation of structured data to improve the performance of code retrieval. However, most of the existing code retrieval systems are combined with code generation tasks such as code generation, code summarization, and code completion to build code retrieval augmented frameworks. The knowledge boundary problem of the language model can be alleviated by retrieving relevant code snippets and documentation from external knowledge bases. Therefore, based on the previous work, we have made the following improvements and extensions: 1) we extend our previous SANTA model into a code assistant (CONAN), which consists of a code structure-aware retriever and a dual-view code representation-based retrieval-augmented generation
model. 2) The dual-view code representation-based retrieval-augmented generation model designs a dual-view code representation mechanism that helps language models better understand code semantics by regarding the code documentation descriptions as prompts. 3) The dual-view code representation-based retrieval-augmented generation model employs the Fusion in Decoder (FID) architecture, which breaks the limitation of the input length of the language model. 4) The code assistant (CONAN) performs well on several code-related tasks including code retrieval, code generation, code summarization and code completion. 5) CONAN can be used as an assistant for the large language models to assist them in finishing various code tasks. \textbf{All codes are available at} \url{https://github.com/NEUIR/CONAN}.}
\footrule

\author{Xinze Li}
\authornote{indicates equal contributions.}
\email{lxzlxz0716@gmail.com}
\author{Hanbin Wang}
\authornotemark[1]
\email{wanghanbinpanda@gmail.com}
\affiliation{%
  \institution{Northeastern University}
  \city{Shenyang}
  \country{China}}

\author{Zhenghao Liu}
\authornote{indicates corresponding author.}
\affiliation{%
  \institution{Northeastern University}
  \city{Shenyang}
  \country{China}}
\email{liuzhenghao@mail.neu.edu.cn}

\author{Shi Yu}
\affiliation{%
  \institution{Tsinghua University}
  \city{Beijing}
  \country{China}}

\author{Shuo Wang}
\affiliation{%
  \institution{Tsinghua University}
  \city{Beijing}
  \country{China}}

\author{Yukun Yan}
\affiliation{%
  \institution{Tsinghua University}
  \city{Beijing}
  \country{China}}

\author{Yukai Fu}
\affiliation{%
  \institution{Chinese Academy of Sciences}
  \city{Shenyang}
  \country{China}}

\author{Yu Gu}
\affiliation{%
  \institution{Northeastern University}
  \city{Shenyang}
  \country{China}}

\author{Ge Yu}
\affiliation{%
  \institution{Northeastern University}
  \city{Shenyang}
  \country{China}}

\renewcommand{\shortauthors}{Li et al.}

\begin{abstract}

Pretrained language models have shown strong effectiveness in code-related tasks, such as code retrieval, code generation, code summarization, and code completion tasks. In this paper, we propose \textbf{CO}de assista\textbf{N}t vi\textbf{A} retrieval-augme\textbf{N}ted language model (CONAN), which aims to build a code assistant by mimicking the knowledge-seeking behaviors of humans during coding. Specifically, it consists of a code structure aware retriever (CONAN-R) and a dual-view code representation-based retrieval-augmented generation model (CONAN-G). CONAN-R pretrains CodeT5 using Code-Documentation Alignment and Masked Entity Prediction tasks to make language models code structure-aware and learn effective representations for code snippets and documentation. Then CONAN-G designs a dual-view code representation mechanism for implementing a retrieval-augmented code generation model. CONAN-G regards the code documentation descriptions as prompts, which help language models better understand the code semantics. Our experiments show that CONAN achieves convincing performance on different code generation tasks and significantly outperforms previous retrieval augmented code generation models. Our further analyses show that CONAN learns tailored representations for both code snippets and documentation by aligning code-documentation data pairs and capturing structural semantics by masking and predicting entities in the code data. Additionally, the retrieved code snippets and documentation provide necessary information from both program language and natural language to assist the code generation process. CONAN can also be used as an assistant for Large Language Models (LLMs), providing LLMs with external knowledge in shorter code document lengths to improve their effectiveness on various code tasks. It shows the ability of CONAN to extract necessary information and help filter out the noise from retrieved code documents.

\end{abstract}






\maketitle

\section{Introduction}
In recent years, the code pertaining technologies~\cite{feng2020codebert,guo2020graphcodebert,wang2021codet5} have shown promissing effectiveness in code related tasks~\cite{DBLP:conf/nips/LuGRHSBCDJTLZSZ21,lu2021codexglue}, such as code generation~\cite{lu2021codexglue,parvez2021retrieval,guo2023retrievalaugmented}, code summarization~\cite{parvez2021retrieval,wang2021codet5,wang2023codet5} and code completion~\cite{lu2022reacc,zhang2023repocoder}. This convincing generation effectiveness allows code developers to understand, modify, and write code more efficiently, making it possible to build an effective code assistant.

\begin{figure}[t] \centering
    \includegraphics[width=0.8\textwidth]{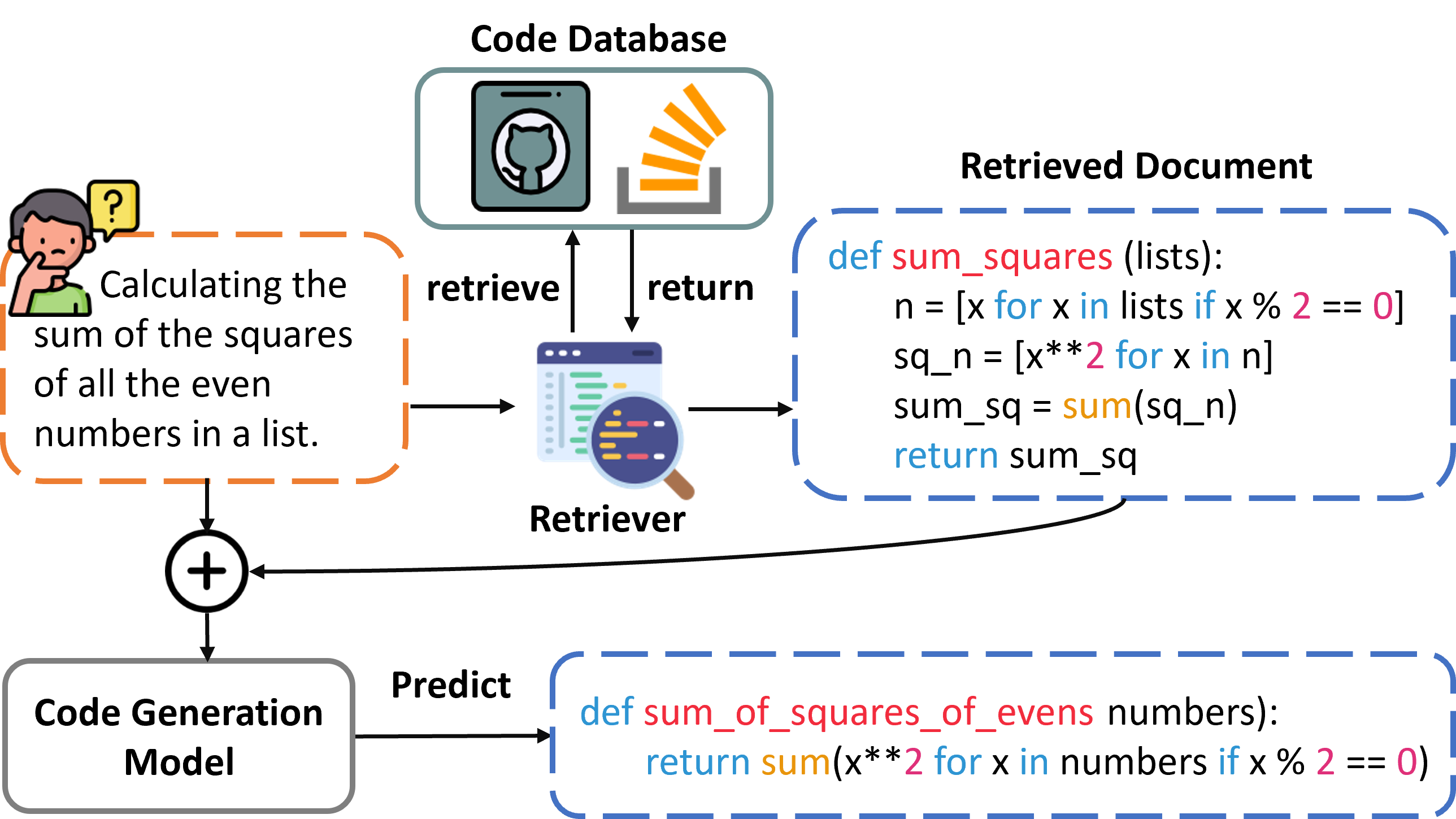}
    \caption{The Motivation of Building A Code Assistant via the Retrieval-Augmented Code Generation Model.} \label{fig:intro}
\end{figure}

As shown in Figure~\ref{fig:intro}, even though the pretraining technique improves the effectiveness of language models on code-oriented tasks, the code generation and understanding ability of language models is limited by the knowledge boundary of input program language (PL) or natural language (NL) which can result in them generating incorrect or unsatisfactory code~\cite{jiang2023active,luo2023sail}. Similarly, in real software development scenarios, many professional software engineers also encounter challenging tasks that are beyond their capabilities and knowledge. Whenever this happens, software engineers usually seek related information from the question-answering forums or the code repository, such as StackOverflow and GitHub, facilitating them understand and write code~\cite{brandt2010example,sadowski2015developers}. These software engineers not only refer to the code documentation and solutions but also copy the code segments (a code repository usually contains 7-23\% cloned parts~\cite{svajlenko2015evaluating}) to increase their development productivity and accelerate software development~\cite{li2013help,roy2008empirical,baker2007finding}.

Inspired by the above scenario, many works have begun to mimic the retrieval and generation behaviors of software engineers during the coding process to build the retrieval augmented model~\cite{lu2022reacc,parvez2021retrieval,shrivastava2023repofusion,liu2021retrievalaugmented,shapkin2023entity,zhou2022docprompting} to improve the performance of language models in the code-related tasks. They utilize different knowledge sources of codes to benefit the code-related tasks~\cite{liao2023context}, \textit{e.g.} using related code segments~\cite{lu2022reacc,parvez2021retrieval}, code documentations~\cite{zhou2022docprompting} or external entities~\cite{shapkin2023entity} to improve the quality of generated codes. These models employ BM25 or DPR~\cite{karpukhin2020dense} to retrieve related code segments and incorporate external coding knowledge by directly concatenating external information~\cite{lu2022reacc,parvez2021retrieval}. The code segments are usually long, making the work only model two code segments while completing the codes~\cite{lu2022reacc}. Nevertheless, the code retrieval process inevitably introduces additional noise. In this case, it is crucial to alleviate the noise of retrieved code segments in building the retrieval augmented code generation model.

In this paper, we propose \textbf{CO}de Assista\textbf{N}t vi\textbf{A} Retrieval-Augme\textbf{N}ted Language Model (CONAN) to build a unified framework and serve code-related tasks, such as the code generation, code summarization, and code completion tasks. CONAN consists of a code structure-aware retriever (CONAN-R) and a dual-view code representation-based retrieval-augmented generation model (CONAN-G), which are designed to alleviate the noise of retrieved code segments. Specifically, CONAN-R reduces the noise of retrieved code knowledge by conducting more accurate retrieval results. Following our previous work~\cite{li2023structure}, CONAN-R designs two pretraining tasks, Code-Documentation Alignment (CDA) and Masked Entity Prediction (MEP) to continuously train CodeT5~\cite{wang2021codet5}. These tasks teach the language model to learn more effective representations for both code segments and documentation for retrieval. The Code-Documentation Alignment (CDA) task contrastively trains PLMs to align matched code-documentation pairs in the embedding space, which better represents codes by bridging the modality gap between program language and natural language. The Masked Entity Prediction (MEP) task masks entities in codes and trains PLMs to fill in the masked parts, which helps to capture semantics from code. Then, to fully use the code knowledge provided by CONAN-R, CONAN-G follows previous work~\cite{shrivastava2023repofusion,zhou2022docprompting} and employs the Fusion-in-Decoder (FID) architecture~\cite{fid} to break the max length limitation of existing language models, making it possible to incorporate multiple retrieved code segments during generation. Besides, CONAN-G regards the code documentation as a gist, stimulates language models to capture more critical semantics from code structures using the code documentation, and alleviates the effect of noise from long code segments. Moreover, CONAN can be used as an assistant to provide Large Language Models (LLMs) with necessary code knowledge. Specifically, CONAN can retrieve relevant code documents from external knowledge databases and further summarize and denoise these retrieved code documents to shorter yet higher-quality documents, which in turn assists the LLMs in completing code-related tasks.

Our experiments show that both the retrieval module and generation module of CONAN achieve convincing performance in code-related tasks, such as code generation, code summarizing, and code completion, providing a promising way to build a code assistant. The effectiveness of CONAN mainly derives from more accurate code retrieval (CONAN-R) and the dual-view code representation-based retrieval-augmented generation model (CONAN-G). On one hand, our further analyses show that CONAN-R achieves state-of-the-art on code retrieval tasks and shows strong zero-shot ability, which can supply more informative code segments for CONAN-G. By aligning structured and unstructured data (Code-Documentation Alignment Task), CONAN-R maps both codes and documentation in one universal embedding space and learns more tailored embeddings for code retrieval. The masked entity prediction task further guides CONAN-R to capture more crucial information for retrieval and better distinguish structured and unstructured data. On the other hand, by leveraging multiple code segments, CONAN-G outperforms baseline code generators and achieves consistent improvements in all code-related tasks. Notably, the code documentations show their effectiveness in guiding generation models to better capture key information from codes, further confirming that the multi-model modeling method can generalize its advantages to the uni-model tasks~\cite{liu2022universal}.
\section{Related Work}
For building a code assistant, lots of work has focused on the Code-related generation tasks~\cite{DBLP:conf/nips/LuGRHSBCDJTLZSZ21}, which include code generation~\cite{lu2021codexglue,parvez2021retrieval,guo2023retrievalaugmented}, code summarization~\cite{parvez2021retrieval,wang2021codet5,wang2023codet5} and code completion~\cite{lu2022reacc,zhang2023repocoder}. Recent work mainly focuses on pretraining language models to deal with code-related generation tasks, facilitating the code development~\cite{li2013help,roy2008empirical,baker2007finding}. To mimic the knowledge-seeking behavior of software engineers during coding, lots of work~\cite{lu2022reacc,parvez2021retrieval,shrivastava2023repofusion,liu2021retrievalaugmented,shapkin2023entity,zhou2022docprompting} focuses on building the retrieval augmented model to improve the performance of language models in the code-related tasks. They utilize different information of codes to further improve the code-related tasks~\cite{liao2023context}.

\textbf{Code-Oriented Language Models.} The code-oriented pretrained language models (PLMs) utilize code corpora for pretraining and design different training strategies to make the language model conduct a deeper understanding of code semantics, such as code syntax, semantics, and idiomatic constructs~\cite{feng2020codebert,ahmad2021unified,Zan2022CERTCP}. CodeBERT uses replaced token detection~\cite{clark2020electra} and masked language modeling~\cite{devlin2019bert} to learn the lexical semantics of structured data~\cite{DBLP:conf/nips/LuGRHSBCDJTLZSZ21}. 
DOBF~\cite{roziere2021dobf} further considers the characteristics of code-related tasks and replaces class, function, and variable names with special tokens.
CodeT5~\cite{wang2021codet5} not only employs the span mask strategy~\cite{raffel2020exploring} but also masks the identifiers in codes to teach T5~\cite{raffel2020exploring} to generate these identifiers, which helps better distinguish and comprehend the identifier information in code-related tasks. Additionally, some researchers leverage multi-modal data such as code, comment, and abstract syntax trees (AST) to pretrain models, enhancing the model's understanding of code, natural language, code structure, and other related information~\cite{li2022coderetriever,guo2022unixcoder}. Recently, Large Language Models (LLMs), such as Llama~\cite{touvron2023llama} and ChatGPT~\cite{chatgpt}, have demonstrated their ability in many code tasks, such as code understanding and code generation. To further improve the ability of LLMs on code tasks, many researchers focus on Continuously pretraining LLMs on large amounts of code pretraining data to enable them to learn sufficient code knowledge. CodeQwen1.5~\cite{codeqwen1.5} is initialized from Qwen1.5 and trained on 3 trillion tokens of code data to make it with strong code generation capabilities. Codellama~\cite{roziere2023code} is pretrained based on Llama2~\cite{touvron2023llama} with a total of 500 billion of generic and code data.

\textbf{Code Retrieval.} The code retrieval models~\cite{li2022coderetriever,li2023structure} usually employ the dense retrieval architecture for searching code segments~\cite{Yu2021FewShotCD,karpukhin2020dense,xiong2020dense,li2021more}. These models encode queries and codes using PLMs~\cite{devlin2019bert,liu2019roberta,raffel2020exploring}, map them in an embedding space for retrieval, and then conduct KNN search in the embedding space~\cite{johnson2019billion}. The query and code encoders are usually contrastively optimized to guarantee the retrieval effectiveness and the negatives are sampled from inbatch training documents, BM25 retrieved documents, and hard negatives~\citep{karpukhin2020dense,xiong2020approximate}.

Leaning more effective representations with PLMs is crucial for dense retrieval~\cite{cocondenser,luan2020sparsedense}, thus several continuous training models are proposed. They usually employ mask language modeling to train PLMs on structured data and help to memorize the semantic knowledge using model parameters~\cite{wang2021codet5,feng2020codebert,roziere2021dobf}. Nevertheless, the mask language modeling~\cite{devlin2019bert} may not sufficiently train PLMs to represent texts and show less effectiveness in text matching tasks~\cite{chen2021exploring,gao2019representation,li2020sentence,reimers2019sentence,li2020sentence}. The recent development of sentence representation learning methods has achieved convincing results~\cite{fang2020cert,yan2021consert}. The work first constructs sentence pairs using back-translation~\cite{fang2020cert}, some easy deformation operations~\cite{wu2020clear}, original sequence cropping~\cite{meng2021coco} or adding dropout noise~\cite{gao2021simcse}. Then they contrastively train PLMs to learn sentence representations that can be used to distinguish the matched sentence pairs with similar semantics. Furthermore, some work also considers the characteristics of codes during pretraining. CodeRetriever~\cite{li2022coderetriever} pretrains PLMs to learn more tailored representations for codes with the unimodal and bimodal contrastive losses, which encourages the model to push codes with similar functionality closer and align the matched code and text in the embedding space. We start from CodeT5 and propose code structure-aware pretraining~\cite{li2023structure} to pretrains language models structure-aware. code structure-aware pretraining designs the Code-Documentation Alignment (CDA) and Masked Entity Prediction (MEP) tasks for pretraining, which teach models to distinguish matched structured data for unstructured texts and ask language models to fill in the masked entities, respectively.

\textbf{Retrieval-Augmented Code Generation.} Code PLMs generate or complete codes based on input natural language descriptions or code snippets~\cite{lu2021codexglue,ahmad2021unified,lu2022reacc}. However, existing code PLMs usually face the knowledge boundary problem of input~\cite{jiang2023active,luo2023sail}, which stimulates researchers to focus more on searching different knowledge for enhancing the code generation performance, \textit{e.g.} using related code snippets~\cite{lu2022reacc,parvez2021retrieval}, code documentations~\cite{zhou2022docprompting} or external entities~\cite{shapkin2023entity} to improve the quality of generated codes and learning necessary information from surrounding context to better understand a repository~\cite{shrivastava2023repofusion}. Even though the work leverages different kinds of knowledge for code generation, they incorporate external coding knowledge by directly concatenating external information~\cite{lu2022reacc,parvez2021retrieval} or leverages the fusion-in-decoder (FID) architecture~\cite{shrivastava2023repofusion,zhou2022docprompting}.

The external knowledge sources seem effective in enhancing the capabilities of generating more accurate code snippets, summaries, and completions. Specifically, REDCODER~\cite{parvez2021retrieval} retrieves relevant code snippets or summaries from a retrieval database using the DPR model~\cite{karpukhin2020dense} and then provides them as a supplement to improve the code generation and summarization performance. ReACC~\cite{lu2022reacc} focuses on the code completion task. It first utilizes the unfinished code as a query and then retrieves a similar code snippet that is completed using the lexical retrieval model. Then the unfinished code and completed code are concatenated and fed into the code generation model. SKCODER~\cite{li2023skcoder} is a sketch-based code generation approach, which extracts a code sketch from the retrieved similar code and further edits the sketch into the target code based on the input description. These existing methods still face challenges in fully using the retrieval information in the generation models. It is evident that the retrieval model returns lots of noise, which limits the effectiveness of code retrieval augmented models. Thus effectively searching and utilizing more related code context as auxiliary information is an ongoing research direction of code PLMs.

\begin{figure*}[t] \centering
    \includegraphics[width=0.95\textwidth]{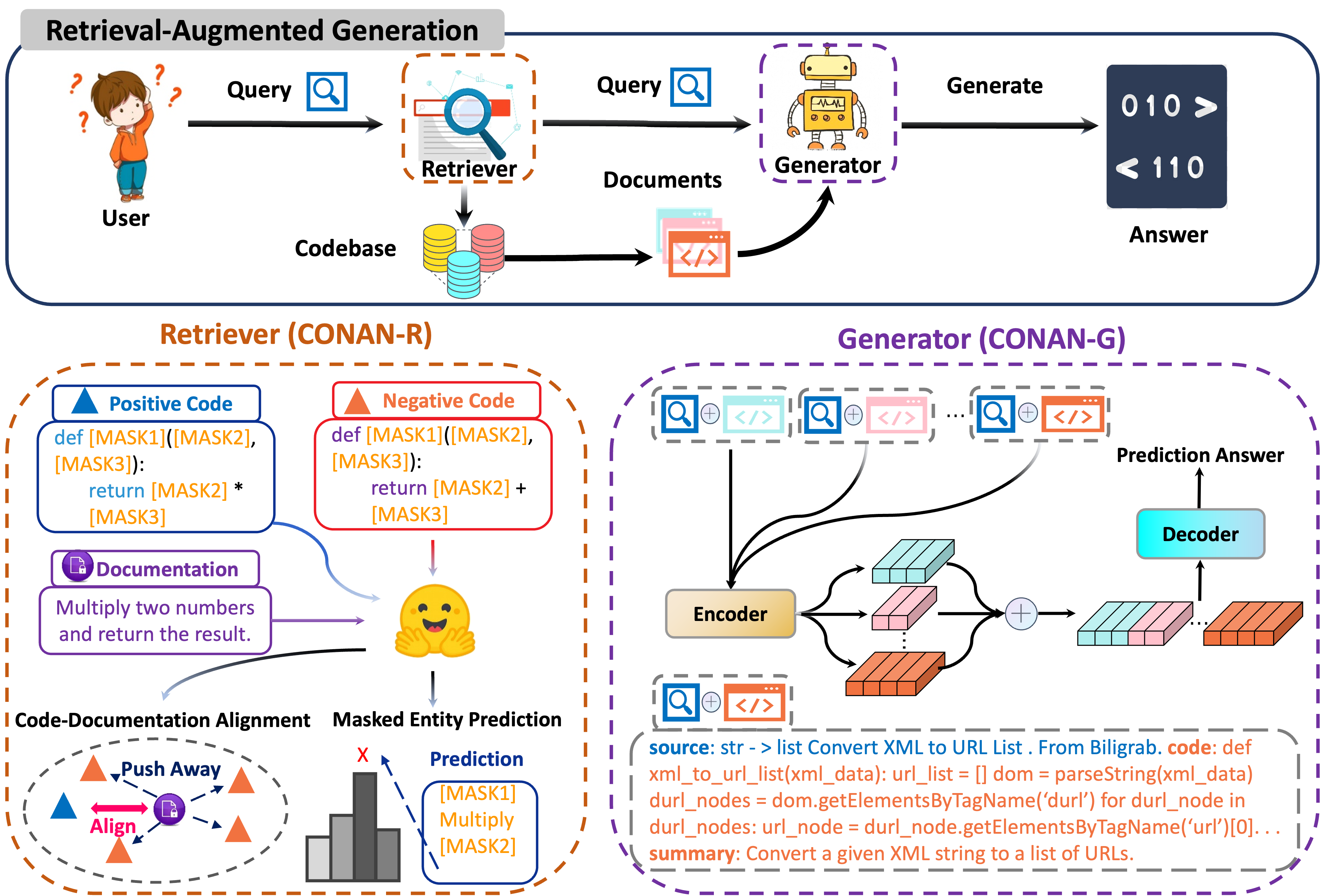}
    \caption{The Architecture of \textbf{CO}de Assista\textbf{N}t vi\textbf{A} Retrieval-Augme\textbf{N}ted Language Model (CONAN). CONAN consists of a code structure-aware retriever (CONAN-R) and a dual-view code representation mechanism (CONAN-G). We employ Code-Documentation Alignment (CDA) and Masked Entity Prediction (MEP) methods for CONAN-R pretraining. CONAN-G is implemented with the Fusion-in-Decoder (FID) architecture.} \label{fig:model}
\end{figure*}

\section{Methodology}
In this section, we introduce the \textbf{CO}de Assista\textbf{N}t vi\textbf{A} Retrieval-Augme\textbf{N}ted Language Model (CONAN) (Figure~\ref{fig:model}). We first introduce the preliminary of the retrieval augmented generation framework (Sec.~\ref{sec:rag}). Then, we describe the retrieval module (CONAN-R) and generation module (CONAN-G) in Sec.~\ref{sec:retrieval} and Sec.~\ref{sec:generate}, respectively. CONAN-R pretrains pretrained language models (PLMs) to better capture the structure semantics of codes and conduct better code representations. CONAN-G utilizes the code documentation description as a gist to better understand code semantics. Finally, we utilize the CONAN model to generate code knowledge and aid LLMs for generating (Sec.~\ref{sec:plug-in}).

\subsection{Preliminary of the Retrieval Augmented Code Generation Framework}\label{sec:rag}
To deal with code generation, completion, and summarization tasks, CONAN regards the code function descriptions, unfinished codes, and code segments as queries $q$ and then retrieves related code documents $D$ as external knowledge to facilitate generating more accurate codes and summarizations. $d \in D$ and $d$ consists of the code snippet $d_\text{code}$ and the code documentation $d_\text{doc}$.

The retrieval augmented generation framework~\cite{fid,guu2020retrieval,lewis2020retrieval} includes a code retriever and a generation model, aiming to retrieve useful information for the generation model and utilize external knowledge to improve the generation accuracy. We utilize CodeT5~\cite{wang2021codet5,wang2023codet5} as backbone PLM and then implement the retrieval and generation modules.

\textbf{Retrieval Module (CONAN-R).} Existing retrieval augmented models usually leverage dense retrievers to conduct efficient search~\cite{fid,guu2020retrieval,lewis2020retrieval}. For the given query $q$, dense retrieval models aim to retrieve related code documents from the external code knowledge corpus. They encode the query $q$ and document $d$ and map them in an embedding space for retrieval. Following the previous work~\cite{li2023structure}, we use CodeT5 to encode the query $q$ and the document $d$ as low dimensional representations $h^q$ and $h^d$, using the representation of the first token from the decoder:
\begin{equation}
h^q = \text{CodeT5}(q); h^d = \text{CodeT5}(d).
\end{equation}
Then we conduct KNN search by calculating the similarity score $f(q, d)$ between the dense representations $h^q$ and $h^d$ of query $q$ and document $d$:
\begin{equation}
f(q,d) = sim(h^q,h^d),
\end{equation}
where $sim$ is the dot product function to calculate the relevance between query $q$ and document $d$. The top-$N$ retrieved documents that are most similar to the query are denoted as $D=\{d^1, d^2, ..., d^N\}$. Specifically, we utilize the code snippet $d_\text{code}$ to represent the document $d$ in the code generation task and use the code documentation $d_\text{doc}$ to represent the document $d$ in the code completion and code summarization tasks.

\textbf{Generation Module (CONAN-G).} To generate the code or natural language sequence $t$, the generative module (CONAN-R) leverages the top-$N$ retrieved documents $D=\{d^1, d^2,...,d^N\}$ from the retrieval model to facilitate the generation process.

To fully use the information from retrieved code documents $D$, we follow previous work~\cite{shrivastava2023repofusion,zhou2022docprompting} and employ the fusion-in-decoder (FID) architecture~\cite{fid} to break the max length boundary of PLMs. The $j$-th token $t_j$ of the generated sequence $t$ can be generated according to the probability $P(t_j|q, D, t_{1,...,j-1})$:
\begin{equation}\label{eq:fid}
P(t_j|q, D, t_{1,...,j-1}) = \text{FID} (q, D, t_{1,...,j-1}).
\end{equation}
Finally, the codes or summarization results can be generated.

\subsection{CONAN-R with Structure Aware Pretraining}\label{sec:retrieval}
To learn a tailored embedding space for code retrieval, CONAN-R finetunes the representations of query and document by minimizing the loss $\mathcal{L}_{\text{CONAN-R}}$:
\begin{equation}
 \mathcal{L}_{\text{CONAN-R}} = -\log\frac{e^{f(q,d^+)}}{e^{f(q,d^+)} + \sum_{d^-\in D^-}{e^{f(q,d^-)}}},
\end{equation}
where $d^+$ is relevant to the given query $q$. $D^-$ is the collection of irrelevant code documents, which are sampled from inbatch negatives~\cite{karpukhin2020dense}. Existing language models are usually pretrained on unstructured natural languages with masked language modeling~\cite{devlin2019bert,liu2019roberta}. Nevertheless, these models struggle to better understand the semantics represented by data structures, which limits the effectiveness of language models in representing code documents for retrieval~\cite{feng2020codebert,wang2021codet5}.

For CONAN, we follow the previous work~\cite{li2023structure} and continuously pretrain the CodeT5 to learn more tailored embedding space for codes using two structure-aware pretraining tasks: code-documentation alignment and masked entity prediction. Through code structure-aware pretraining, pretrained language models further capture the structural semantics of the code and better learn the representation of code snippet, which can retrieve high-quality multi-view knowledge for the generator model.


\textbf{Code-Documentation Alignment (CDA).} The Code-Documentation Alignment task teaches language models to optimize the embedding space by aligning code snippet with documentation. 

For each code snippet $d_\text{code}$, the document $d$ usually contains the code documentation $d_\text{doc}$ that has the same semantics as $d_\text{code}$. We can utilize the underlying semantic connections between these natural language based code documentation $d_\text{doc}$ and code snippet $d_\text{code}$ to perform alignment to train the language model to better represent code snippet. 

Specifically, we can use CodeT5 to encode the code documentation $d_\text{doc}$ and code snippet $d_\text{code}$ as $h_\text{code}^d$ and $h_\text{doc}^d$, respectively, calculate the similarity score $f(d_\text{doc},d_\text{code})$ between $d_\text{doc}$ and $d_\text{code}$, and then continuously train language models using the contrastive loss $\mathcal{L}_{\text{CDA}}$:
\begin{equation}\label{eq:sta}
\begin{aligned}
 &\mathcal{L}_{\text{CDA}} = -\log \frac{e^{f(d_\text{doc},d_\text{code}^+)}}{e^{f(d_\text{doc},d_\text{code}^+)}+\sum_{d_\text{code}^- \in D_\text{code}^-} e^{f(d_\text{doc},d_\text{code}^-)}},
\end{aligned}
\end{equation}
where $d_\text{code}^+$ is relevant code snippet to the given $d_\text{doc}$. $D_\text{code}^-$ consists of the irrelevant code snippet $d_\text{code}^-$ sampled from in-batch negatives. The contrastive training method can bridge the semantic gap between code snippets and natural language documentation and map them in one universal embedding space, benefiting learning representations of multi-modal text data~\cite{liu2022universal}.

\textbf{Masked Entity Prediction (MEP).} The masked entity prediction guides the language models to better understand the semantics of code snippets by recovering masked entities.
We mask entities for continuous training language models instead of using the random masking strategy in mask language modeling~\cite{devlin2019bert,raffel2020exploring}.

As shown in previous work~\cite{sciavolino2021simple,zhang2019ernie}, entity semantics show strong effectiveness in learning text data representations during retrieval. Thus, we first recognize mentioned entities that appeared in the document $X_d=\{x_1, \text{ent}_1, x_2, \text{ent}_2, ..., \text{ent}_n\}$ and mask them as the input for T5 encoder module:
\begin{equation}
X_d^\text{mask} = \{x_1, \text{mask}_1, x_2, \text{mask}_2, ..., x_n\},
\end{equation}
where $\text{mask}_i$ is a special token to denote the $i$-th masked span. We replace the same entity with the same special token. Then 
 we continuously train T5 to recover these masked entities using the following loss function:
\begin{equation}\label{eq:mep}
\mathcal{L}_{\text{MEP}} = \sum_{j=1}^k -\log P(Y_d (t_j)| X_d^\text{mask}, Y_d (t_{1,...,j-1})),
\end{equation}
where $Y_d (t_j)$ denotes the $j$-th token in the sequence $Y_d$. And $Y_d = \{ \text{mask}_1, \text{ent}_1, ..., \text{mask}_n, \text{ent}_n\}$ denotes the ground truth sequence that contains masked entities. During training, we optimize the language model to fill up masked spans and better capture entity semantics by picking up the necessary information from contexts to recover the masked entities, understanding the structure semantics of code snippet~\cite{ye2020coreferential}.

\subsection{CONAN-G with Dual-View Code Representation}\label{sec:generate}
As shown in Eq.~\ref{eq:fid}, we use the Fusion-in-Decoder (FID) architecture to fully use the information of retrieved code documents $D$ to benefit the code generation process. To optimize the parameters of CONAN-G, we train the model with the following loss function $\mathcal{L}_{\text{CONAN-G}}$:
\begin{equation}\label{eq:train_lm}
 \mathcal{L}_{\text{CONAN-G}} = \sum_{j=1}^m -\log P(t_j^*|q, D, t_{1,...,j-1}) = \sum_{j=1}^m -\log \text{FID} (q, D, t_{1,...,j-1}),
\end{equation}
where $t_j^*$ denotes the $j$-th golden token of target sequence $t$ and the sequence contains $m$ tokens. The FID decoder module is inherited from CodeT5-Decoder. Then it uses the encoded representations $\text{Enc} (q, D)$ of query and retrieved documents and the embeddings $\{e_t^1, e_t^2, ..., e_t^{j-1}\}$ of the tokens $\{t_1, t_2, ..., t_{j-1}\}$ to calculate the generation probability of the next token $t_j$:
\begin{equation}
\text{FID} (q, D, t_{1,...,j-1})=\text{CodeT5-Decoder}(\text{Enc} (q, D), \{e_t^1, e_t^2, ..., e_t^{j-1}\}),
\end{equation}
where the Enc function separately encodes the query $q$ and code documents $D$ using the CodeT5-Encoder:
\begin{equation}
\text{Enc} (q, D) =\text{CodeT5-Encoder}(d^1\oplus q) \oplus, ..., \oplus \text{CodeT5-Encoder}(d^N\oplus q),
\end{equation}
where $\oplus$ is the concatenation operation. 

For the document $d^i$, we can represent it using the code documentation $d^i_\text{doc}$ and code segment $d^i_\text{code}$, which describe the function of the document in natural language and program language, respectively. In CONAN, we propose a dual-view code representation method and simply concatenate the text sequences of the code documentation $d^i_\text{doc}$ and code segment $d^i_\text{code}$ to better represent the document:
\begin{equation}
\text{CodeT5-Encoder}(d^i\oplus q) = \text{CodeT5-Encoder}(d^i_\text{doc}\oplus d^i_\text{code}\oplus q).
\end{equation}
The dual-view code representation method regards the code documentation $d^i_\text{doc}$ as a kind of gist to help PLMs better understand code semantics. It thrives on the strong language understanding ability of pretrained language models and then utilizes functional instruction to capture crucial information from the code structure.

\subsection{Assisting LLMs for Code Related Tasks Using CONAN}\label{sec:plug-in}
Beside dealing with different code related tasks, CONAN can be also used as a code assistant to aid LLMs for generating codes. In this case, CONAN aims to retrieve code segments and then extract necessary knowledge from retrieved code segments, enabling LLMs to access the external code knowledge and filter out the noise from retrieved contents. 

Specifically, for a query $q$, CONAN first uses the query $q$ to retrieve the relevant code documents $D$ from the whole database $\widetilde{D}$:
\begin{equation}
D = \text{CONAN-R}(q, \widetilde{D}),
\end{equation}
where \text{CONAN-R} is the retrieval module of CONAN. Then, we use CONAN-G to extract the code knowledge from these retrieved code documents $D$ by generating a new code document $d^*$:
\begin{equation}
d^* = \text{CONAN-G}(q, D),
\end{equation}
where $d^*$ represents the summarization results, code snippet and code segments for code summarization task, code completion task and code generation task, respectively.
Finally, we use $d^*$ as the augmented knowledge for the code LLMs to assist LLMs to generate outputs $y$ during solving different code-related tasks:
\begin{equation}
y = \text{LLM}(d^* \oplus q),
\end{equation}
where we regard the generated code knowledge $d^*$ as the context and concatenate it with the given query $q$ to support the inference of LLMs.
\section{Experimental Methodology}
In this section, we describe the datasets, retrieval databases, evaluation metrics, baselines, and implementation details of our experiments.

\subsection{Dataset}
In this subsection, we introduce the datasets used in pretraining CONAN-R and code-related generation tasks.

\begin{figure}[t]
    \centering
    \subfigure[Constructed Pretrained Data Pairs.] { \includegraphics[width=0.49\linewidth]{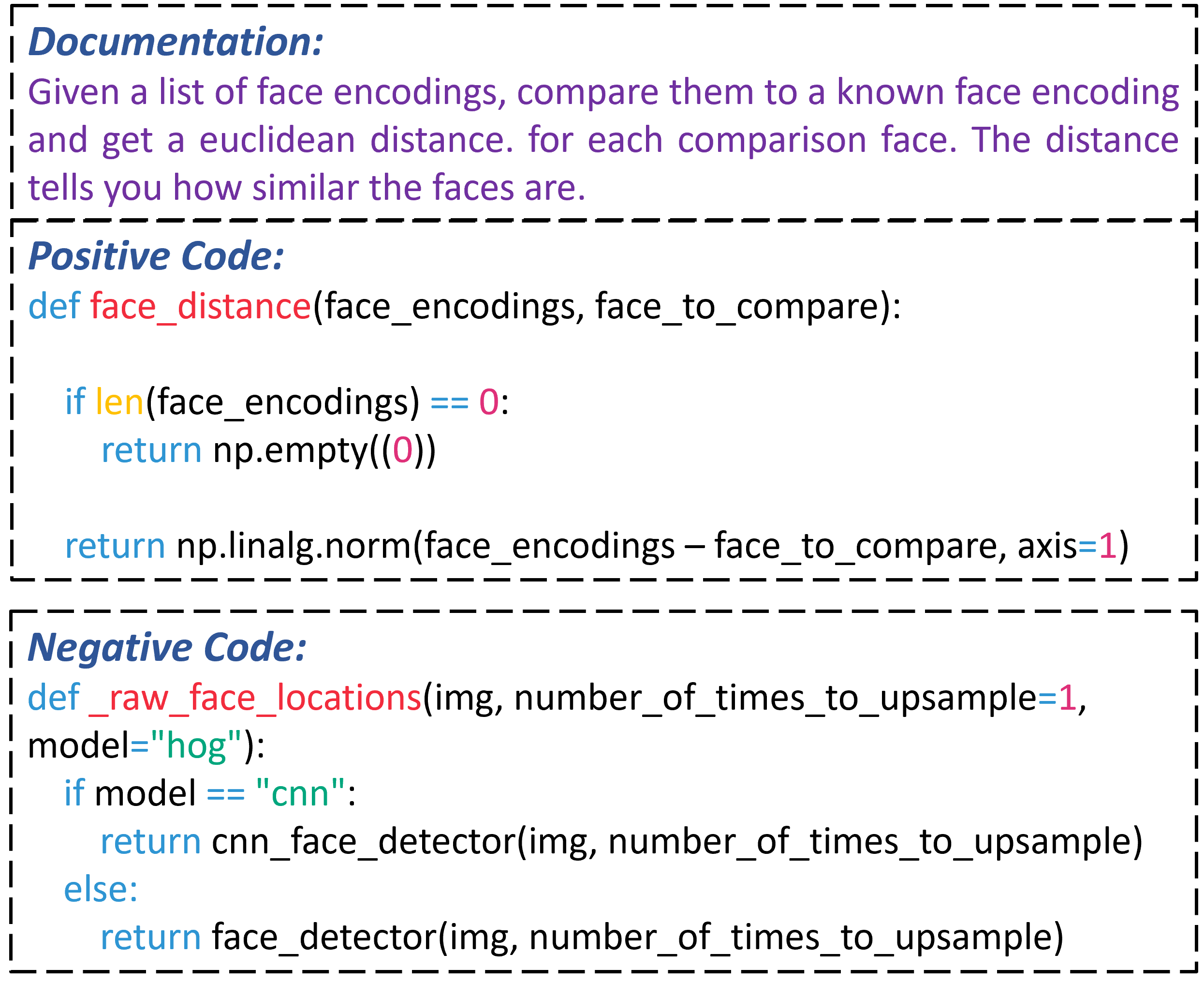}}
    \label{fig:construct_data_pos_neg}
    \subfigure[Identify Entity.] {\includegraphics[width=0.49\linewidth]{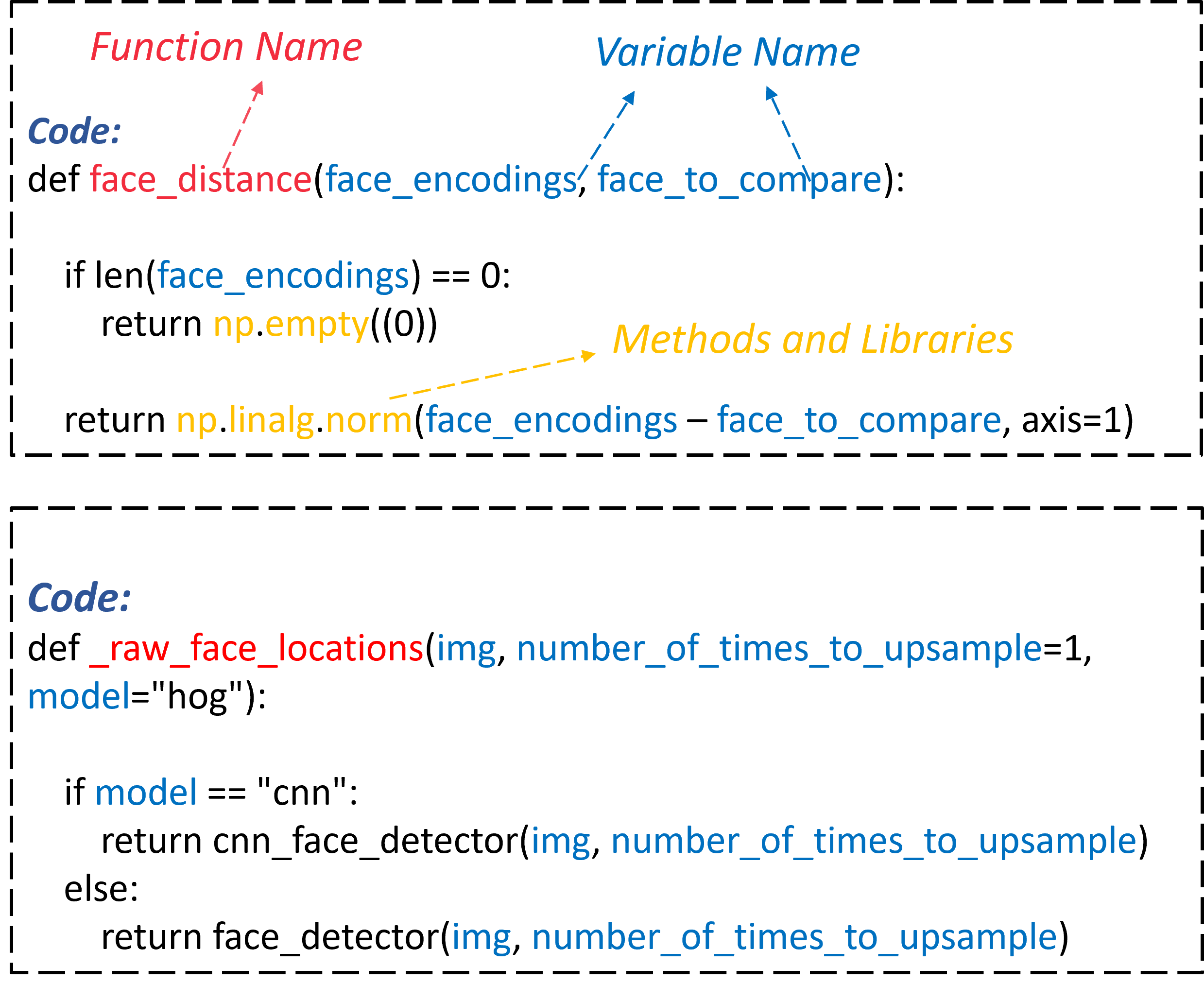}}
    \label{fig:construct data identity}

    \caption{Examples of the Pretraining Data for CONAN-R. All entities of different functions are annotated with different colors in Figure 3 (b).}
    \label{fig:construct_data}
\end{figure}

     
\begin{table}
\begin{center}
\caption{Data Statistics of Pretraining Data. ``Entities'' denotes the proportion of identified entities in the code data.}
\label{tab:pretrain-data}
\small
\begin{tabular}{l| r r}
\hline
{\textbf{Task}} & {\textbf{Positive Pairs}} & {\textbf{Entities}} \\ 
\hline
Python & 429,596 & 28.6\%  \\
PHP & 514,127 &  17.8\% \\
Go & 317,824 & 17.1\% \\ 
Java & 454,433 & 24.4\%  \\
JavaScript & 122,682 & 15.4\% \\
Ruby  & 48,790 & 28.8\% \\
\hline
\end{tabular}
\end{center}
\end{table}
\textbf{Retrieval.}
Firstly, we present the pretraining data CodeSearchNet, which is used to pretrain our CONAN-R model. The data statistics are shown in Table~\ref{tab:pretrain-data}.

During pretraining CONAN-R, we use the CodeSearchNet in experiments. As shown in Figure~\ref{fig:construct_data}, we present an example to show how to construct the code-documentation pairs for pretraining. The code snippets have corresponding code documentations, which describe the purpose and function of these code snippets. As shown in Figure~\ref{fig:construct_data} (a), the code documentation and its corresponding code snippet are regarded as a training pair. Then we regard the documentation as a query and use inbatch negatives to optimize T5 for code retrieval pretraining.
Additionally, as shown in Figure~\ref{fig:construct_data} (b), we follow~\citet{wang2021codet5} and regard code identifiers such as variables, function names, external libraries, and methods as entities. Then we replace the same entities with the same special tokens and ask CONAN-R to generate these masked entities (Eq.~\ref{eq:mep}). These special tokens come from the vocabulary of T5, such as \{\text{<extra\_id\_0>}, \text{<extra\_id\_1>}, ..., \text{<extra\_id\_99>} \}. BytesIO and tree\_sitter\footnote{\url{https://github.com/tree-sitter/tree-sitter}} are utilized to identify entities in Python and other programming languages, respectively. The proportions of identified entities in pretraining data are shown in Table~\ref{tab:pretrain-data}.

\begin{table*}[t]
\centering
\small
\caption{Dataset Statistics for Code-related Generation Tasks. $\left|\textbf{Code}\right|$ and $\left|\textbf{Doc}\right|$ represent the average lengths of code and documentation in the dataset for code generation and summarization tasks, respectively. In code completion, they respectively represent the average length of incomplete code and the length of code to be completed. All the code and documentation lengths are calculated before tokenization.}
\label{tab:statistic_gen}
\begin{tabular}{l|l|l|r|r|r|r|r}
\hline
\textbf{Task}                       & \textbf{Dataset}               & \textbf{Lang} & \multicolumn{1}{c|}{\textbf{Train}} & \multicolumn{1}{c|}{\textbf{Dev}} & \multicolumn{1}{c|}{\textbf{Test}} & \multicolumn{1}{c|}{$\left|\textbf{Code}\right|$} & \multicolumn{1}{c}{$\left|\textbf{Doc}\right|$} \\ \hline
\multirow{5}{*}{Code Generation}    & \multirow{2}{*}{CgCSN} & Python        & 251,820         & 13,914        & 14,918         & 99              & 14                          \\
                                    &                                & Java          & 164,923         & 5,183         & 10,955         & 97              & 12                          \\  
                                    & Concode                        & Java          & 100,000         & 2,000         & 2,000          & 27              & 72
                                    \\  
                                    & HumanEval                        & Python          & -         & -         & 164          & 66              & 156
                                    \\ 
                                    & MBPP                        & Python          & -        & -         & 500          & 17              & 78
                                    \\ \hline
\multirow{2}{*}{Code Summarization} & \multirow{2}{*}{CsCSN} & Python        & 251,820         & 13,914        & 14,918         & 99              & 14                          \\
                                    &                                & Java          & 164,923         & 5,183         & 10,955         & 97              & 12                          \\ \hline
\multirow{2}{*}{Code Completion}    & PY150                          & Python        & 68,589          & 3,825         & 10,000         & 497             & 7                           \\
                                    & JavaCorpus                     & Java          & 11,774          & 4,993         & 3,000          & 639             & 11                          \\ \hline
\end{tabular}
\end{table*}
\textbf{Code-Related Generation.}
We describe the datasets used to evaluate the generation effectiveness of CONAN-G, including code generation, code summarization, and code completion datasets. The data statistics are shown in Table~\ref{tab:statistic_gen}.

\textit{Code Generation.} The code generation task aims to generate the code snippets according to the given code documentation description. We first utilize the Concode~\cite{iyer2018mapping} and CgCSN~\cite{parvez2021retrieval} datasets to evaluate the code generation ability of different models. For the Concode~\cite{iyer2018mapping} dataset, the input not only includes a natural language description but also encompasses the class environment. This dataset is particularly challenging because the desired code can vary significantly based on the functionality provided by the class. Besides, we follow previous work~\cite{parvez2021retrieval} and use the CgCSN dataset in the code summarization task. CgCSN dataset is filtered from CodeSearchNet~\cite{husain2020codesearchnet}. Additionally, we use HumanEval~\cite{chen2021evaluatinglargelanguagemodels} and MBPP~\cite{austin2021programsynthesislargelanguage} to further test the ability of CONAN to assist the LLMs to generate codes. This dataset focuses more on evaluating the code generation ability of models whether they can pass the test cases.

\textit{Code Summarization.}
The code summarization task is a reversed task of code generation, which generates a code documentation description according to the code snippet. In the code summarization task, we use the CsCSN dataset in our experiments, which is filtered from CodeSearchNet~\cite{husain2020codesearchnet} and includes two programming languages, Python and Java. Some examples that the code cannot be parsed into an abstract syntax tree have been removed. This preprocessing method ensures that the dataset keeps a high quality.

\textit{Code Completion.}
The code completion task targets completing unfinished codes. In this experiment, we employ the PY150~\cite{raychev2016probabilistic} and JavaCorpus~\cite{allamanis2013mining} datasets to evaluate the generation performance. In our experiments, we specifically focus on line-level code completion, which auto-complete a line based on the provided incomplete code snippet.

\begin{table}[t]
\centering
\caption{The Statistics of the Retrieval Database. ``Paired'' indicates that the code snippet/documentation in the retrieval database has corresponding code documentation/snippet. ``Unpaired'' indicates that the code snippet/documentation in the retrieval database does not have corresponding code documentation/snippet. ``$*$'' signifies that code candidates include not only Java code but also class environments.}
\label{tab:retrieval_database}
\small
\begin{tabular}{l|l|l|c|c|c}
\hline
\textbf{Database}                           & \textbf{Lang} & \textbf{Task} & \textbf{Total} & \textbf{Paired} & \textbf{Unpaired}  \\ \hline
\multirow{3}{*}{Code Snippets}    & Python    &CgCSN-Python  & 1.2M          & 696K  & 507K          \\
                                            & Java   & CgCSN-Java        & 1.6M          & 1.1M     &0.5M        \\
                                            & Java$*$  & Concode       & 104K          & 104K  & -            \\ \hline
\multirow{2}{*}{Code Documentation}  & Python    & CsCSN-Python \& PY150    & 1.1M          & 267K  & 833K           \\
                                          & Java    & CsCSN-Java \& JavaCorpus      & 1.1M          & 197K  & 903K           \\ \hline
\end{tabular}
\end{table}
\textbf{Retrieval Databases.}
We follow~\citet{parvez2021retrieval} to construct two retrieval databases, namely the 1) \textit{code snippets retrieval database} and 2) \textit{code documentation retrieval database}. In our experiments, we exclude the target code/documentation from the retrieval database to prevent information leakage from the generation dataset. The statistical information is shown in Table~\ref{tab:retrieval_database}.

\textit{Code Snippet.}
The code snippet corpus is built based on CodeSearchNet~\cite{husain2020codesearchnet}. After deduplication, the code retrieval database contains 1.2 million Python functions and 1.6 million Java functions. Approximately 40\% of these functions are paired with corresponding natural language descriptions. For Concode dataset, we merge its training and validation sets to create a code retrieval database, making all code snippets have corresponding natural language descriptions.

\textit{Code Documentation.}
The code documentation corpus is built based on the combination of high-quality natural language documentation from both CodeSearchNet and CCSD~\cite{liu2021retrievalaugmented}. After deduplication, we retained 1.1 million code documentations, with approximately 20\% of them containing corresponding Java and Python code snippets.

\begin{table}[t]
\centering
\caption{Data Statistics of Code Retrieval Datasets. Two datasets, Adv and CodeSearchNet, are used in our experiments.}
\label{tab:finetune}
\small
\begin{tabular}{l|l|r|r|r}
\hline
\textbf{Dataset}                     & \textbf{Language} & \multicolumn{1}{c|}{\textbf{Train}} & \multicolumn{1}{c|}{\textbf{Dev}} & \multicolumn{1}{c}{\textbf{Test}} \\ \hline
\textbf{Adv}                         & Python        & 251,820                              & 9,604                              & 19,210                             \\ \hline
\multirow{6}{*}{\textbf{CodeSearch}} & Python        & 251,820                              & 13,914                             & 14,918                             \\
                                     & PHP           & 241,241                              & 12,982                             & 14,014                             \\
                                     & Go            & 167,288                              & 7,325                              & 8,122                              \\
                                     & Java          & 164,923                              & 5,183                              & 10,955                             \\
                                     & JavaScript    & 58,025                               & 3,885                              & 3,291                              \\
                                     & Ruby          & 24,927                               & 1,400                              & 1,261                              \\ \hline
\end{tabular}
\end{table}
\subsection{Evaluation Metrics}
In this subsection, we describe the evaluation metrics to test the retrieval and generation performance of CONAN.

To evaluate the retrieval performance of CONAN-R, we finetune CONAN-R and then evaluate its retrieval effectiveness using two code retrieval datasets, Adv~\cite{DBLP:conf/nips/LuGRHSBCDJTLZSZ21} and CodeSearch, which are filtered out from CodeSearchNet dataset ~\cite{husain2020codesearchnet}. The data statistics of the finetuning data are shown in Table~\ref{tab:finetune}. CodeSearch consists of code retrieval tasks on six programming languages, including Ruby, Javascript, Go, Python, Java, and PHP, which can evaluate the model's performance across a diverse set of programming languages. We use MRR@100 to evaluate the performance of the structure-aware code retriever CONAN-R, which is the same as the previous work~\cite{li2022coderetriever,lu2021codexglue,guo2022unixcoder}.

To evaluate the generation performance of CONAN-G, we utilize different evaluation metrics for different tasks. We utilize the corpus level BLEU~\cite{papineni2002bleu}, CodeBLEU (CBLEU)~\cite{ren2020codebleu}, and Pass@$k$ as the evaluation metrics for code generation. We use the smoothed BLEU-4~\cite{lin-och-2004-orange} as the evaluation metric for code summarization. For code completion tasks, we adopt exact match accuracy (EM) and edit similarity (ES) to evaluate line-level code completion.

\subsection{Baselines}

In this subsection, we describe the baselines used in our experiments. We evaluate CONAN against several state-of-the-art code-related generation models and code retrieval models. All baseline models are categorized into two groups: 1) retrieval models, and 2) generation models.

\textbf{Code retrieval models.} We compare CONAN-R with three typical and task-specific code retrieval models to demonstrate its retrieval effectiveness, CodeBERT, CodeT5, and CodeRetriever~\cite{li2022coderetriever}. CodeRetriever is the state-of-the-art code retrieval models, which continuously trains GraphCodeBERT~\cite{guo2020graphcodebert} with unimodal and bimodal contrastive training losses.

\textbf{Code generation models.} The code generation models can be grouped into four categories, including retrieval models, pretrained language models (PLMs), PLM w. RAG models, and large language models (LLMs).

\textit{Retrieval models.} Following prior work~\cite{parvez2021retrieval}, we use the top-ranked code snippets/documentation from the retrieval results as the prediction results. We consider the term-based sparse retriever BM25~\cite{robertson2009probabilistic} and three dense retrievers, CodeBERT~\cite{feng2020codebert}, GraphCodeBERT~\cite{guo2020graphcodebert} and SCODE-R~\cite{parvez2021retrieval} as baseline models. CodeBERT inherits the BERT architecture and is trained on code corpus using both mask language modeling and replaced token detection. GraphCodeBERT is pretrained by modeling the data flow graph of the source code. SCODE-R builds upon the DPR~\cite{karpukhin2020dense} model and uses CodeBERT and GraphCodeBERT as the code and summary encoders.

\textit{PLMs.} The generative model produces the output based on the original input, without external information. CodeGPT and CodeGPT-adapted~\cite{lu2021codexglue} are both decoder-only transformer models pretrained on Python and Java datasets from CodeSearchNet. The former is trained from scratch, while the latter is obtained by continuously training from GPT-2. PLBART~\cite{ahmad2021unified} is a seq2seq model that is capable of performing a broad spectrum of program and language understanding and generation tasks. CodeT5~\cite{wang2021codet5} bases on the T5 architecture. It not only has excellent code understanding capabilities but also possesses strong code generation abilities. UniXcoder~\cite{guo2022unixcoder} is a unified cross-modal pretrained model that leverages multimodal data (i.e. code comment and AST) to pretrain code representations. 

\textit{PLM w. RAG.} These models utilize different retrieval techniques to gather relevant information from a retrieval database, which is then used to guide and enhance the generative models. REDCODER~\cite{parvez2021retrieval} is a framework that retrieves relevant code snippets or documentation from a retrieval database and provides them as a supplement to code generation or summarization models. ReACC~\cite{lu2022reacc} is a retrieval-augmented code completion framework. It adopts a stage-wise approach that combines a source code retriever and an auto-regressive language model for programming language. ReACC-bm25, ReACC-dense, and ReACC-hybrid are three implementations of ReACC, each of which employs a different retriever.

\textit{LLMs.} Moreover, we consider CONAN as an assistant to help LLMs solve different code tasks. Specifically, we use CONAN-R to retrieve relevant code snippets and documentation from external knowledge bases. And then we use CONAN-G to summarize and denoise the retrieved contents to obtain higher-quality knowledge, which is used to assist code LLMs. In our experiments, we use Deepseek-Coder-6.7b-Instruct (DSCoder-6.7b-Ins)~\cite{guo2024deepseekcoderlargelanguagemodel} and CodeQwen1.5-7B-Chat (CQwen1.5-7B-Chat)~\cite{codeqwen1.5} as code LLMs.

\subsection{Implementation Details}
In this subsection, we describe the experimental details of CONAN.

We initialize CONAN-R with CodeT5-base. During the structure-aware pretraining, we set the learning rate as 1e-4 and the training epoch as 10. During finetuning, we train CONAN-R using inbatch negatives and hard negatives. For CodeSearch and Adv datasets, we set the learning rate as 2e-5 and 1e-5, respectively, and set batch size and epoch as 128 and 12. We use inbatch negatives plus one hard negative for finetuning and the hard negative is randomly sampled from the top-100 retrieved negative codes by the finetuned CONAN-R (Inbatch) model. For Concode and CsCSN, we set the learning rate as 2e-5, while for CgCSN, we set the learning rate as 1e-5. For all three datasets, we set batch size and epoch as 64 and 10. We use the Adam optimizer and set the warmup proportion as 0.1. All models are implemented with OpenMatch~\cite{10.1145/3539618.3591813}.

We initialize CONAN-G based on CodeT5-base with the Fusion-in-Decoder (FID) architecture. CONAN-G utilizes a dual-view code representation method, which regards the code documentation description as a gist. However, some documents do not contain the code documentation. Thus, for these instances, we directly use the code snippets to represent the code documents. During training CONAN-G for all code-related generation tasks, we use the retrieved top-5 code snippets and documentation as external knowledge. On the Concode dataset, we set the learning rate as 1e-4. For other datasets, we set the learning rate as 5e-5. For all datasets, we set the batch size as 1, set max epoch as 1, use the AdamW optimizer, and configure the warmup steps as 1,000. All models are implemented with PyTorch and Huggingface transformers~\cite{wolf2020huggingfaces}. When evaluating LLMs on HumanEval and MBPP, we set the temperature to 0.2 and the maximum generation length to 512 tokens.

\begin{table*}[t]
\centering
\caption{Evaluation Results of Code Generation and Code Summarization on Concode and CsCSN Datasets. CsCSN-P and CsCSN-J represent the subsets of the CsCSN dataset to evaluate the code summarization effectiveness in Python and Java programming languages. The baseline results of PLMs setting are reported from PLBART~\cite{ahmad2021unified} and REDCODER~\cite{parvez2021retrieval}.}
\label{tab:Concode_CsCSN}
\small
\begin{tabular}{ll|ccc|ccc}
\hline
  \multicolumn{1}{l|}{\multirow{3}{*}{\textbf{Setting}}}      &\multirow{3}{*}{\textbf{Models}} & \multicolumn{3}{c|}{\textbf{Code Generation}} & \multicolumn{2}{c}{\textbf{Code Summarization}} \\ 
                                                                                         \multicolumn{1}{l|}{}   & &\multicolumn{3}{c|}{\textbf{Concode}}             & \multicolumn{1}{c}{\textbf{CsCSN-P}}  & \multicolumn{1}{c}{\textbf{CsCSN-J}}              \\ \cline{3-7}
                                                                                      \multicolumn{1}{l|}{}  &            & EM             & BLEU           & CBLEU       & BLEU             & BLEU               \\ \hline
\multicolumn{1}{l|}{\multirow{6}{*}{\begin{tabular}[c]{@{}l@{}}Retrieval\\  Models\end{tabular}}}                 
& BM25            & 0              & 20.3           & 23.7          & 1.9               & 1.8                      \\
\multicolumn{1}{l|}{}                                                           & CodeBERT~\cite{feng2020codebert}        & 0              & 27.7          & 41.4          & 11.6              & 12.1            \\
\multicolumn{1}{l|}{}                                                           & CodeT5~\cite{wang2021codet5}        & 0              & 31.1          & 34.9          & 14.6              & 15.7            \\
\multicolumn{1}{l|}{}                                                           & SCODE-R~\cite{parvez2021retrieval}        & 0              & 32.6          & 36.5          & 15.0              & 15.9            \\
\multicolumn{1}{l|}{}                                                                                              & CONAN-R   & 0              & 33.5          & 37.5          & 15.6          & 16.2          \\ \hline
\multicolumn{1}{l|}{\multirow{9}{*}{PLMs}}   
& Seq2Seq~\cite{DBLP:conf/emnlp/LuongPM15}          & 3.1         & 21.3          & 26.4              & 15.9           & 15.1  \\
\multicolumn{1}{l|}{}   
& GPT-2~\cite{radford2019language}        & 17.4             & 25.4          & 29.7          & -              & -                 \\

\multicolumn{1}{l|}{}   
& CodeGPT-2~\cite{DBLP:conf/nips/LuGRHSBCDJTLZSZ21}       & 18.3             & 28.7           & 32.7         & -              & -               \\

\multicolumn{1}{l|}{}   
& CodeGPT-adapted~\cite{DBLP:conf/nips/LuGRHSBCDJTLZSZ21}      & 20.1              & 32.8          & 36.0          & -              & -            \\

\multicolumn{1}{l|}{}   
& CodeBERT~\cite{feng2020codebert}      & 18.0            & 28.7          & 31.4          & 19.1              & 17.7                  \\
\multicolumn{1}{l|}{}                                                            & GraphCodeBERT~\cite{guo2020graphcodebert}  & 18.7             & 33.4          & 35.9          & 18.0             & 17.9                \\
\multicolumn{1}{l|}{}                                                            & PLBART~\cite{ahmad2021unified} & 18.6           & 36.7           & 38.5         & 19.3              & 18.5             \\
\multicolumn{1}{l|}{}                                                            & UniXcoder~\cite{guo2022unixcoder}         & 22.6             & 38.2            & -         & 19.3              & -                   \\
\multicolumn{1}{l|}{}                                                                                              & CodeT5 (Ours)~\cite{wang2021codet5}          & 22.2           & 39.6           & 43.8        & 20.4              &20.5                \\ \hline
\multicolumn{1}{l|}{\multirow{4}{*}{\begin{tabular}[c]{@{}l@{}}PLM \\w. RAG\end{tabular}}} & BM25 + PLBART~\cite{parvez2021retrieval}  & 21.4           & 40.2           & 41.8          & 19.6           & 19.7               \\
\multicolumn{1}{l|}{}                                                                                              & REDCODER~\cite{parvez2021retrieval}       &  23.4          & 41.6           & 43.4          & 21.0           & 22.9              \\
\multicolumn{1}{l|}{}                                                                                              & REDCODER-EXT~\cite{parvez2021retrieval}   & 23.3           & 42.5          & 43.4        & 20.9         & 22.9               \\

\multicolumn{1}{l|}{}                                                                                              & CONAN     &  23.1 & 42.8 & 45.1 & 23.5 & 26.5 \\ 
\hline
\multicolumn{1}{l|}{\multirow{6}{*}{\begin{tabular}[c]{@{}l@{}}LLMs\end{tabular}}} 
 &DSCoder-6.7b-Ins &0	 &7.7	 &12.9 &5.1 &4.4\\
  \multicolumn{1}{l|}{}&DSCoder-6.7b-Ins + CONAN-R  
&14.1	&12.8	&38.2 &18.8	&19.8

	\\

 \multicolumn{1}{l|}{}  &DSCoder-6.7b-Ins + CONAN &\textbf{24.2}	&42.4	&\textbf{45.8}
&\textbf{23.9}	&\textbf{27.1}
	\\ 
  \multicolumn{1}{l|}{} & CQwen1.5-7B-Chat
&0	&8.5	&16.5 &3.2	 &4.6

 \\
  \multicolumn{1}{l|}{} & CQwen1.5-7B-Chat
 + CONAN-R &18.0 &30.5 &38.5 &4.8	&5.2
 \\
 
  \multicolumn{1}{l|}{} & CQwen1.5-7B-Chat
 + CONAN   &\textbf{24.2}	 &\textbf{43.1}	 &44.8 &19.7	&24.8

 \\\hline

\end{tabular}
\end{table*}
\section{Evaluation Result}
In this section, we first explore the performance of CONAN on different code-related generation tasks and verify the denoising effect of CONAN. Then, we conduct ablation studies to show the effectiveness of different modules in CONAN. The effectiveness of structure aware retriever pretraining and dual-view code representation-based retrieval-augmented generation
models are presented. Finally, case studies are shown.

\subsection{Overall Performance}
In this subsection, we show the overall performance of CONAN on code-related generation tasks, including code generation, code summarization, and code completion.

\begin{table*}[t]
\centering
\caption{Code Generation Results on the CgCSN Dataset. The baseline results of PLMs setting are reported from REDCODER~\cite{parvez2021retrieval}.}
\label{tab:gen_csn}
\small
\begin{tabular}{l|l|ccc|ccc}
\hline
\multirow{2}{*}{\textbf{Setting}}  &\multirow{2}{*}{\textbf{Model}}                                                                                                & \multicolumn{3}{c|}{\textbf{Python}}             & \multicolumn{3}{c}{\textbf{Java}}                \\ \cline{3-8}
                                                                                          &             & EM             & BLEU           & CBLEU       & EM             & BLEU           & CBLEU       \\ \hline
\multicolumn{1}{l|}{\multirow{6}{*}{\begin{tabular}[c]{@{}l@{}}Retrieval\\ Models\end{tabular}}}                 & BM25            & 0              & 6.6           & 13.5          & 0              & 4.9            & 16.0             \\
\multicolumn{1}{l|}{}                                                                                              & CodeBERT ~\cite{feng2020codebert}       & 0              & 19.8          & 20.1          & 0              & 21.3          & 22.8          \\

\multicolumn{1}{l|}{}                                                                                              & CodeT5 ~\cite{wang2021codet5}       & 0              & 23.1          & 23.3          & 0              & 26.0         & 27.1          \\
\multicolumn{1}{l|}{}                                                                                              & SCODE-R ~\cite{parvez2021retrieval}       & 0              & 22.8          & 23.9          & 0              & 25.3          & 26.7          \\
\multicolumn{1}{l|}{}                                                                                              & CONAN-R   & 0              & 25.0          & 25.6          & 0              & 28.1          & 31.6          \\ \hline
\multicolumn{1}{l|}{\multirow{5}{*}{PLMs}}                                                                   & CodeBERT ~\cite{feng2020codebert}       & 0              & 4.1          & 10.4          & 0              & 8.4           & 14.5          \\
\multicolumn{1}{l|}{}                                                                                              & GraphCodeBERT ~\cite{guo2020graphcodebert}   & 0              & 4.0           & 10.6          & 0              & 7.9           & 14.5          \\
\multicolumn{1}{l|}{}                                                                                              & CodeGPT-adapted~\cite{DBLP:conf/nips/LuGRHSBCDJTLZSZ21} & 0           & 3.1           & 11.3          & 0              & 7.1            & 14.9           \\
\multicolumn{1}{l|}{}                                                                                              & PLBART ~\cite{ahmad2021unified}         & 0              & 4.9           & 12.0          & 0              & 10.1           & 15.0          \\
\multicolumn{1}{l|}{}                                                                                              & CodeT5 (Ours) ~\cite{wang2021codet5}         & 0           & 6.3           & 14.8          & 0                &12.2                &17.8               \\ \hline
\multicolumn{1}{l|}{\multirow{4}{*}{\begin{tabular}[c]{@{}l@{}}PLM \\w. RAG\end{tabular}}} & BM25 + PLBART~\cite{parvez2021retrieval}   & 0           & 7.0           & 13.9          & 0.1            & 11.4          & 15.5          \\
\multicolumn{1}{l|}{}                                                                                              & REDCODER~\cite{parvez2021retrieval}          & 8.9           & 22.7          & 28.9          & 9.0           & 26.9          & 31.2          \\
\multicolumn{1}{l|}{}                                                                                              & REDCODER-EXT~\cite{parvez2021retrieval}      & 9.6           & 24.4          & 30.2        & 10.2         & 29.0          & 33.2          \\

\multicolumn{1}{l|}{}                                                                                              & CONAN     & 14.6 & 32.9 & 37.3 & 17.2 & 37.7 & 45.4 \\\hline

\multicolumn{1}{l|}{\multirow{6}{*}{\begin{tabular}[c]{@{}l@{}}LLMs\end{tabular}}} 
 &DSCoder-6.7b-Ins & 0	&4.2&	10.2
 & 0	&6.4	&13.3 \\
  \multicolumn{1}{l|}{}&DSCoder-6.7b-Ins + CONAN-R &2.2	&5.9	&16.6

&4.3	&13.7	&24.7\\

 \multicolumn{1}{l|}{}  &DSCoder-6.7b-Ins + CONAN & \textbf{20.5}	&\textbf{33.2}	& \textbf{37.3 }&21.4	&\textbf{38.1}	&45.5\\ 
  \multicolumn{1}{l|}{} & CQwen1.5-7B-Chat
&0	&5.7	&9.8	&0	&9.8	&20.5
 \\
  \multicolumn{1}{l|}{} & CQwen1.5-7B-Chat
 + CONAN-R &1.34	 &7.9	 &15.6	 &2.6	 &10.5	 &21.3
\\
 
  \multicolumn{1}{l|}{} & CQwen1.5-7B-Chat
 + CONAN  &19.4	&32.9	&37.1	&\textbf{22.5}	&\textbf{38.1}	&\textbf{46.2}
 \\\hline
\end{tabular}
\end{table*}

As shown in Table~\ref{tab:Concode_CsCSN}, we first show the code generation and code summarization performance of CONAN on the Concode and CsCSN datasets. 
The code generation and code summarization tasks generate code snippets and code documentation descriptions according to the code documentation descriptions and code snippets, which aims to estimate the code understanding and generation ability. Overall, CONAN achieves the highest BLEU and CBLEU scores among all baseline models and also surpasses the state-of-the-art models REDCODER-EXT with an average of approximately 3.1\% and 0.6\% improvements on CsCSN and Concode datasets, respectively. It shows the effectiveness of our CONAN model. 

For the baseline models, the models that are in Retrieval
Models setting even outperform the CodeGPT2 model, demonstrating that the retrieved code snippets and documentation can help to answer the given question. It confirms the crucial roles of retrieved code snippets, which have many overlaps with the ground truth answers. Among all models in Retrieval Models setting, CONAN-R outperforms BM25, CodeBERT, GraphCodeBERT, CodeT5, and SCODE-R in terms of BLEU and CBLEU scores for both Python and Java programming languages of both code generation and summarization tasks. This indicates the effectiveness of CONAN-R in retrieving more relevant code snippets or documentation descriptions for the given queries. The reason for the EM value being 0 is that we filter out the ground truth answers from the retrieval results. We do this because, in real-world scenarios, the ground truth answers are rarely exactly matched with the retrieved code snippets. Thrived on external knowledge, the retrieval augmented model shows much better performance than the vanilla code/summarization generation models. CONAN achieves more than 2\% improvements than our main baseline model CodeT5, which illustrates that CONAN can use the external code knowledge for generating. Besides, CONAN also outperforms all models in PLM
w. RAG setting, showing the effectiveness of the code-aware pretraining method for CONAN-R and the dual-view code representation method for CONAN-G. When utilizing CONAN as an assistant for the code large language models (LLMs setting), CONAN-R can retrieve external knowledge to improve the performance of code LLMs on Concode and CsCSN datasets (DSCoder-6.7b-Ins + CONAN-R and CQwen1.5-7B-Chat + CONAN-R). In addition, when we use CONAN-G to summarize and denoise this retrieved external knowledge, the effectiveness of the code LLMs on Concode and CsCSN is further improved (DSCoder-6.7b-Ins + CONAN and CQwen1.5-7B-Chat + CONAN), achieving an approximately 10\% improvement in EM. This indicates that CONAN can refine the retrieved knowledge to improve the performance of code LLMs by filtering out noise and irrelevant information.

Then, as shown in Table~\ref{tab:gen_csn}, we further evaluate the CONAN model on the CgCSN dataset. The average code length of this dataset is 98, which is much longer than the Concode dataset which is 27. CONAN achieves more significant improvements on the CgCSN dataset than the performance on the Concode dataset (approximately 7\% improvements on CgCSN and 0.6\% improvements on Concode). The improvements demonstrate that CONAN has the ability to better understand the code semantics and fully use the external code knowledge to generate longer codes of the higher-quality, which illustrates the advantages of CONAN in dealing with real-world code generation problems and the possibility of building a code assistant. Besides, utilizing the denoising knowledge of CONAN can further improve the accuracy of code LLMs on code generation tasks.

\begin{table*}
\centering
\caption{Evaluation Results on the Code Completion Task. The baseline results of PLMs setting are reported from ReACC~\cite{lu2022reacc}.}
\label{tab:com}
\small
\begin{tabular}{l|l|cccc}
\hline
\multirow{2}{*}{\textbf{Setting}}    & \multirow{2}{*}{\textbf{Model}}                                                                                          & \multicolumn{2}{c}{\textbf{PY150}} & \multicolumn{2}{c}{\textbf{JavaCorpus}} \\ \cline{3-6}
                                                                                       &             & EM      & ES        & EM        & ES         \\ \hline
\multicolumn{1}{l|}{\multirow{9}{*}{PLMs}}                                                                 & LSTM~\cite{memory2010long}        & 17.93  & 50.05          &  10.30              & 41.55               \\
\multicolumn{1}{l|}{}                                                                                            & Transformer ~\cite{vaswani2017attention}   &  36.65           &  67.51           &  15.33                 & 50.39             \\
\multicolumn{1}{l|}{}                                                                                            & GPT-2 ~\cite{radford2019language}          & 41.73            & 70.60            & 27.50               & 60.36              \\
\multicolumn{1}{l|}{}                                                                                            & CodeGPT~\cite{DBLP:conf/nips/LuGRHSBCDJTLZSZ21}          & 42.18            & 71.23           & 28.23              & 61.81              \\
\multicolumn{1}{l|}{}                                                                                            & CodeGPT-adapted~\cite{DBLP:conf/nips/LuGRHSBCDJTLZSZ21}  & 42.37            & 71.59           & 30.60               & 63.45              \\
\multicolumn{1}{l|}{}                                                                                            & CodeT5-base~\cite{wang2021codet5}     & 36.97            & 67.12           & 24.80               & 58.31              \\
\multicolumn{1}{l|}{}                                                                                            & CodeT5 (Ours)~\cite{wang2021codet5}     & 35.99            & 66.76           & 25.20               & 57.99              \\
\multicolumn{1}{l|}{}                                                                                            & PLBART ~\cite{ahmad2021unified}         & 38.01            & 68.46           & 26.97              & 61.59              \\
\multicolumn{1}{l|}{}                                                                                            & UniXcoder ~\cite{guo2022unixcoder}        & 43.12            & 72.00              & 32.90              & 65.78              \\ \hline
\multicolumn{1}{l|}{\multirow{4}{*}{\begin{tabular}[c]{@{}l@{}}PLM\\ w. RAG\end{tabular}}} 
& ReACC-bm25 ~\cite{lu2022reacc}     & 46.07            & 73.84           & 30.63              & 64.28              \\
\multicolumn{1}{l|}{}                                                                                            & ReACC-dense~\cite{lu2022reacc}       & 45.32            & 73.95           & 30.30               & 64.43              \\
\multicolumn{1}{l|}{}                                                                                            & ReACC-hybrid ~\cite{lu2022reacc}     & \textbf{46.26}   & \textbf{74.41}  & \textbf{30.70}      & \textbf{64.73}     \\

\multicolumn{1}{l|}{}                                                                                            & CONAN     & 40.12            & 69.44           & 26.02              & 62.86              \\ \hline

\multicolumn{1}{l|}{\multirow{6}{*}{\begin{tabular}[c]{@{}l@{}}LLMs \end{tabular}}} 
 & DSCoder-6.7b-Ins & 22.65            & 54.89           & 17.52              & 50.39 \\
 
  \multicolumn{1}{l|}{}& DSCoder-6.7b-Ins + CONAN-R  
	 &  36.67            & 61.90 &    24.59           & 50.40\\

 \multicolumn{1}{l|}{}  & DSCoder-6.7b-Ins + CONAN &44.69	&73.26	&\textbf{29.80}	&\textbf{65.22}
	\\ 
  \multicolumn{1}{l|}{} & CQwen1.5-7B-Chat
& 19.40            & 47.77 &15.55               &42.60 

 \\
  \multicolumn{1}{l|}{} & CQwen1.5-7B-Chat
 + CONAN-R &29.80 & 66.57 & 24.80 & 58.31\\
 
  \multicolumn{1}{l|}{} & CQwen1.5-7B-Chat
 + CONAN   
	&\textbf{45.50}	&\textbf{73.51}	&29.00	&64.33

 \\\hline
\end{tabular}
\end{table*}

Additionally, we evaluate the code completion capability of CONAN on PY150 and Github JavaCorpus, and the experimental results are shown in Table~\ref{tab:com}. In our experiment, CONAN does not outperform the state-of-the-art model ReACC (decoder-only architecture), which mainly lies in the different backbone generation models that they use. However, compared to our main baseline model CodeT5, CONAN achieves more than 3\% improvements on average, which demonstrates that the supplementary retrieval information is helpful. These T5-based models are pretrained with a span denoising training objective and may be not more tailored for code completion tasks than the auto-regressive generation models, which is also observed in previous work~\cite{lu2022reacc,wang2023codet5}. Furthermore, we can observe that using CONAN as an assistant for code LLMs can achieve competitive results compared to ReACC, which validates the effectiveness of utilizing CONAN to summarize and denoise the retrieved knowledge.


\begin{table*}[t]
\centering
\caption{Code Generation Results on the HumanEval and MBPP Datasets. DSCoder and CQwen represent Deepseek-Coder-6.7b-Instruct and CodeQwen1.5-7B-Chat respectively. Top-$k$ stands for selecting the top-ranked $k$ code snippets/documentation from CONAN-R retrieval results as external knowledge to assist in code large language models. The highest results are in \textbf{bold} and the second highest scores are \underline{underlined}.}
\label{tab:humaneval}
\resizebox{\linewidth}{!}{
\begin{tabular}{l|cccc|cccc}
\hline
\textbf{Dataset} & \textbf{DSCoder} & \textbf{w/ Top-1} & \textbf{w/ Top-5} & \textbf{w/ CONAN} & \textbf{CQwen} & \textbf{w/ Top-1} & \textbf{w/ Top-5} & \textbf{w/ CONAN} \\ \hline
HumanEval        & \underline{75.0}             & \textbf{77.4}                             &\textbf{77.4}                             & \textbf{77.4}                     & 77.9              & 78.8                              & \textbf{80.5}                              & \underline{79.3}                      \\
MBPP             & 68.9             & 68.9                             & \textbf{70.2}                             & \underline{69.9}                     & \textbf{71.9}              & 70.2                              & \underline{71.2}                              & 70.9                      \\ \hline
\end{tabular}
}
\end{table*}
Finally, we further evaluate the performance of CONAN as a code LLMs assistant on HumanEval and MBPP datasets. As shown in Table \ref{tab:humaneval}, we observe that using CONAN's denoised knowledge can achieve comparable results to using the Top-$5$ documents retrieved. Moreover, compared to the top-ranked documents, CONAN's denoised results enhance the generation quality of LLMs. This indicates that CONAN possesses the ability to effectively extract relevant information from massive data and denoise them, enabling it to assist LLMs with shorter yet higher-quality texts.

\begin{table*}[t]
\centering
\caption{Ablation Study. We show the effectiveness of the retrieval-augmented generation (RAG) module, the fusion-in-decoder (FID) based dual-view code representation module, and the natural language and program language based code representation method (Dual-View).}
\label{tab:abla_gen}
\resizebox{\linewidth}{!}{
\begin{tabular}{l|cccccc|cc|cccc}
\hline
\multirow{3}{*}{\textbf{Methods}} & \multicolumn{6}{c|}{\textbf{Code Generation}}                                                                                                   & \multicolumn{2}{c|}{\textbf{Code Summarization}}         & \multicolumn{4}{c}{\textbf{Code Completion}}                                                 \\ \cline{2-13} 
                                  & \multicolumn{2}{c|}{\textbf{CgCSN-P}}              & \multicolumn{2}{c|}{\textbf{CgCSN-J}}              & \multicolumn{2}{c|}{\textbf{Concode}} & \multicolumn{1}{c|}{\textbf{CsCSN-P}} & \textbf{CsCSN-J} & \multicolumn{2}{c|}{\textbf{PY150}}                & \multicolumn{2}{c}{\textbf{JavaCorpus}} \\ \cline{2-13}
                              & BLEU          & \multicolumn{1}{c|}{CBLEU}      & BLEU          & \multicolumn{1}{c|}{CBLEU}      & BLEU              & CBLEU          & \multicolumn{1}{c|}{BLEU}           & BLEU           & EM   & \multicolumn{1}{c|}{ES}      & EM        & ES           \\ \hline
CONAN                            & \textbf{32.9} & \multicolumn{1}{c|}{\textbf{37.3}} & \textbf{37.7} & \multicolumn{1}{c|}{\textbf{45.4}} & 42.8              & \textbf{45.1}     & \multicolumn{1}{c|}{\textbf{23.5}}    & \textbf{26.5}    & \textbf{40.1} & \multicolumn{1}{c|}{\textbf{69.4}} & 26.0               & \textbf{62.9}      \\
w/o RAG                           & 6.3          & \multicolumn{1}{c|}{14.8}          & 12.2           & \multicolumn{1}{c|}{17.8}          & 39.6              & 43.8              & \multicolumn{1}{c|}{20.4}             & 20.5             & 36.0          & \multicolumn{1}{c|}{66.8}          & 25.2               & 58.0               \\
w/o FID                           & 27.5          & \multicolumn{1}{c|}{31.7}          & 32.0            & \multicolumn{1}{c|}{36.5}          & 41.9              & 44.7              & \multicolumn{1}{c|}{22.1}             & 23.8             & 39.9          & \multicolumn{1}{c|}{67.4}          & 25.4               & 60.2               \\
w/o Dual-View                       & 30.3          & \multicolumn{1}{c|}{35.1}          & 35.0            & \multicolumn{1}{c|}{41.9}          & \textbf{42.9}     & 44.9              & \multicolumn{1}{c|}{23.4}             & 25.9             & 38.7          & \multicolumn{1}{c|}{68.4}          & \textbf{26.5}      & 60.9               \\ \hline

CodeBERT+CONAN-G                           & 25.2          & \multicolumn{1}{c|}{31.4}          & 29.0            & \multicolumn{1}{c|}{32.6}          & 41.6              & 43.4              & \multicolumn{1}{c|}{21.5}             & 21.9             & 36.7          & \multicolumn{1}{c|}{67.1}          & 25.2               & 59.6               \\
CodeT5+CONAN-G                        & 27.5          & \multicolumn{1}{c|}{33.8}          & 33.6            & \multicolumn{1}{c|}{40.3}          & 42.1     & 44.7              & \multicolumn{1}{c|}{21.9}             & 23.5             & 38.0          & \multicolumn{1}{c|}{67.5}          & 25.7      &61.5                \\ \hline

\end{tabular}}
\end{table*}
\subsection{Ablation Study}
In this subsection, we conduct ablation studies to explore the roles of individual modules of CONAN. 

As shown in Table~\ref{tab:abla_gen}, we study the effectiveness of generation models using different retrieval-augmented methods, including CONAN w/o RAG, CONAN w/o FID and CONAN w/o Dual-View. The CONAN w/o RAG model does not incorporate additional code documents during generation, which is the same as the CodeT5. Then the CONAN w/o FID model keeps the same model architecture with CodeT5 and directly concatenates the top-ranked code documents with queries to augment the code/summarization generation capability. CONAN w/o Dual-View only uses the code snippets to augment the model.

Our experimental results show that the advantages of CONAN mainly derive from the external retrieved code knowledge. Compared with CONAN w/o RAG, CONAN w/o FID achieves about 7.6\% improvements, showing that the external knowledge benefits the code/summarization generation capability of CONAN. Then the Fusion-in-Decoding (FID) model further brings 3\% improvements than CONAN w/o FID, which demonstrates the effectiveness of FID architecture in modeling external retrieved knowledge. The improvements mainly lie in that the FID architecture has the ability to overcome the max length limitation of PLMs, denoise the retrieval results, and fully model the external knowledge, which are also observed in previous work~\cite{fid}.

Then we explore the effectiveness of the dual-view code representation method in CONAN-G. Our dual-view-based code document representation method achieves on average 1.85\%, 0.05\%, and 0.98\% improvements on the code generation, code summarization, and code completion tasks, respectively. The better code generation/completion results demonstrate that the code documentation descriptions indeed help the model better understand the code semantics, making the code generation model better copy and refer to the retrieved code snippets to generate more accurate code results. Our dual-view code representation mechanism shows less effectiveness in the code summarization task. The main reason mainly lies in that only 25\% of the code snippets have corresponding code documentation in the retrieval database. Additionally, we replace CONAN-R with CodeBERT and CodeT5, which have inferior retrieval performance, and observe a decrease in model performance. This demonstrates that CONAN-R can retrieve higher-quality auxiliary information to guide generation, validating the effectiveness of CONAN-R.

\begin{figure}
  \centering
  \subfigure[Code Generation.]{\includegraphics[width=0.32\linewidth]{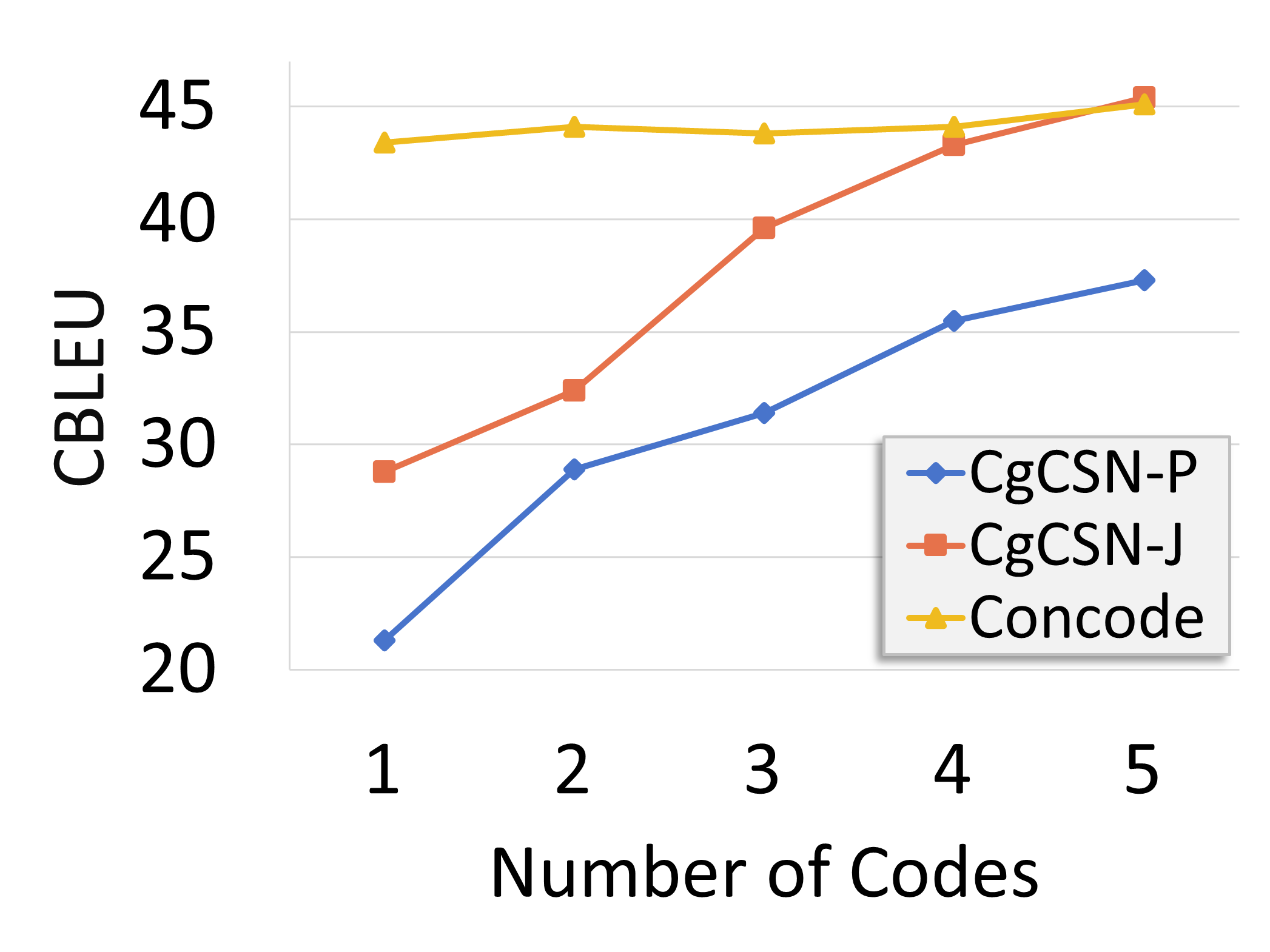}\label{fig:i1}}
  \subfigure[Code Summarization.]{\includegraphics[width=0.32\linewidth]{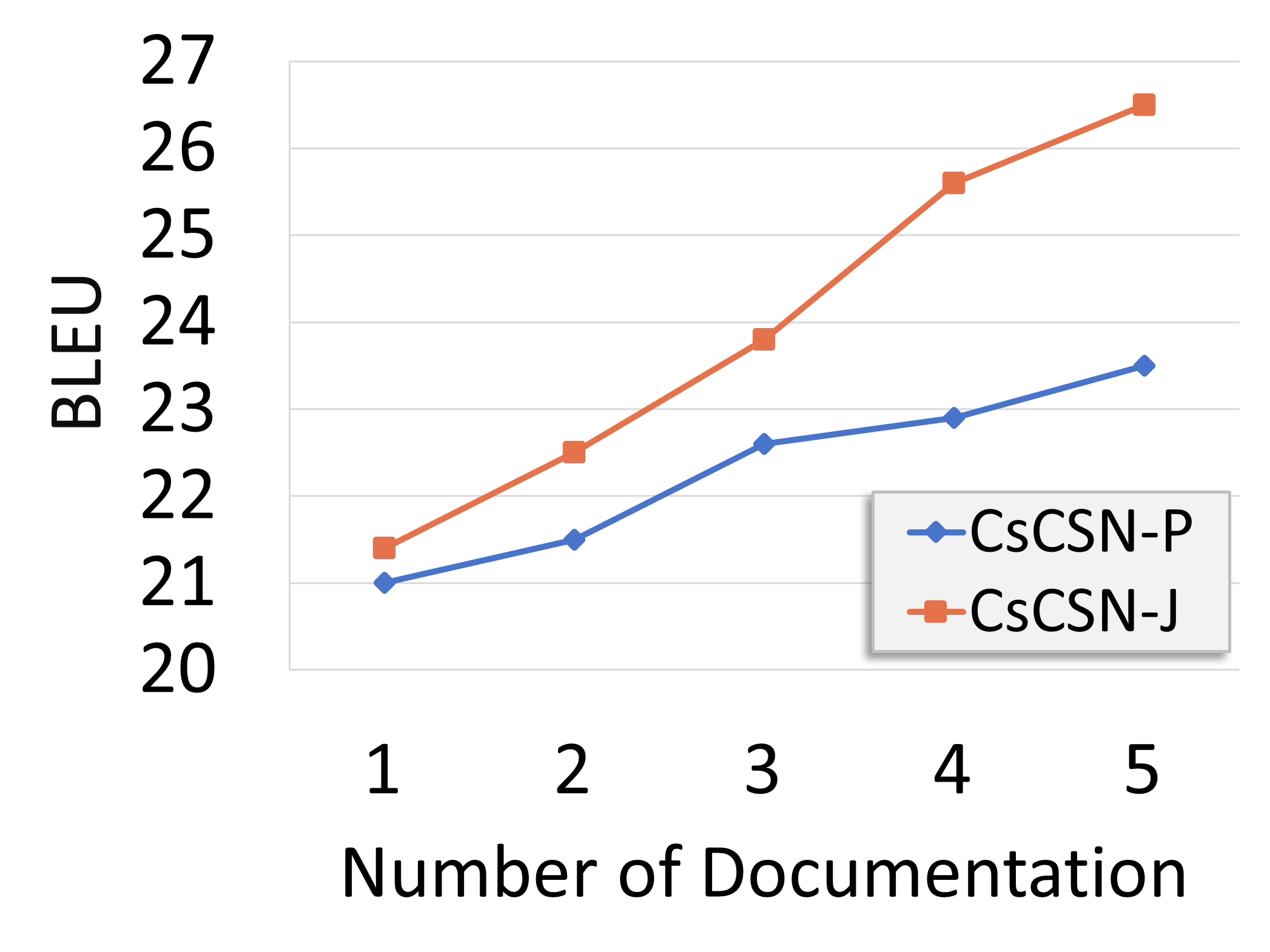}\label{fig:i2}}
  \subfigure[Code Completion.]{\includegraphics[width=0.32\linewidth]{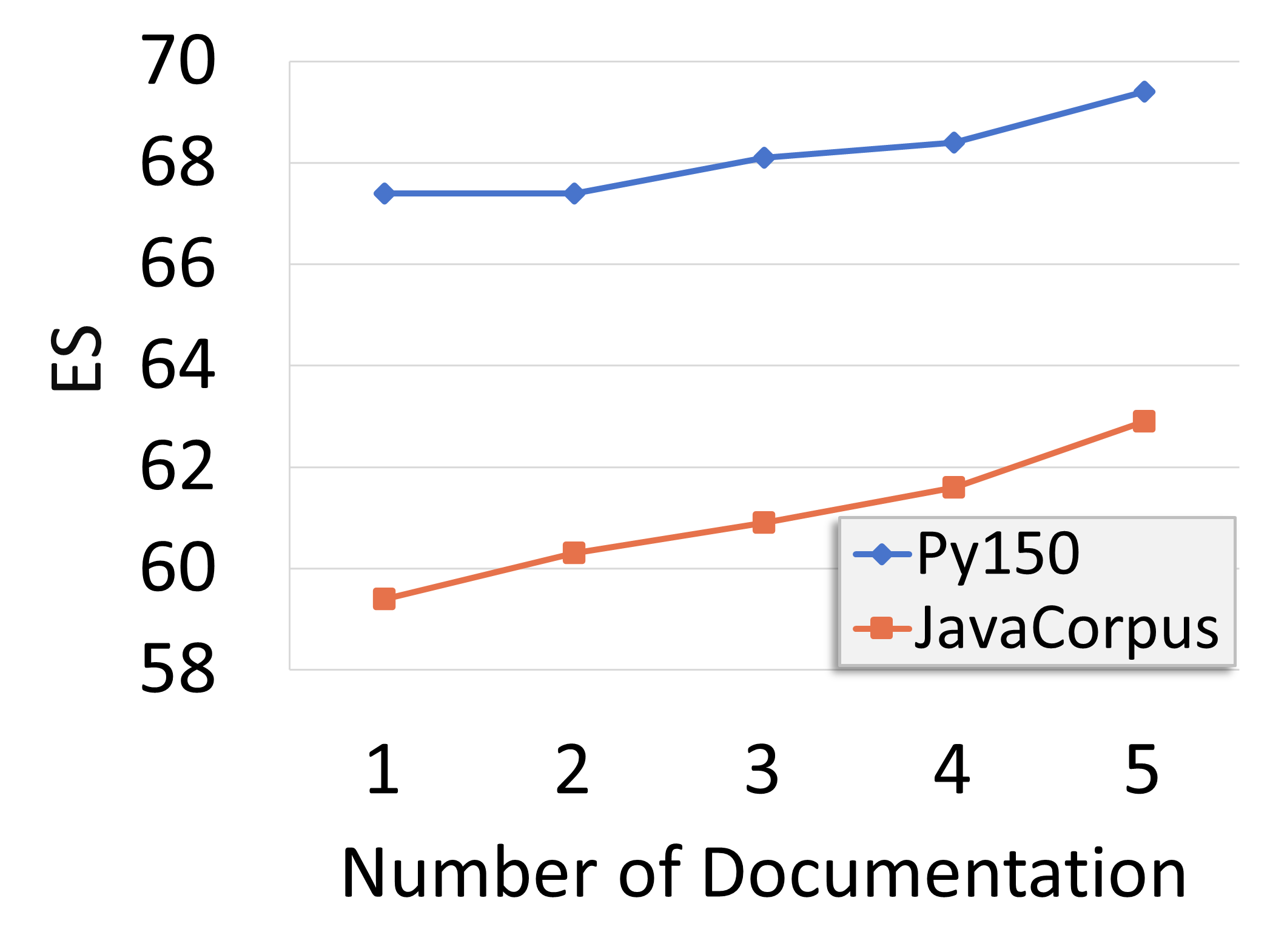}\label{fig:i3}}

  \caption{The impact of the number of retrieved code snippets/documentation on CONAN’s performance.}
  \label{fig:numbers}
\end{figure}
Finally, we explore the impact of the number of retrieved code snippets/documentation on CONAN's performance. As shown in Figure~\ref{fig:numbers}, we observe that increasing the number of retrieved code snippets/documentation leads to a continuous improvement on CONAN's performance in code generation and code summarization. We believe that this is evidence that CONAN excels at combining information from multiple passages.

\begin{figure}[t]
    \centering
    \subfigure[Code Generation.] { \includegraphics[width=0.4\linewidth]{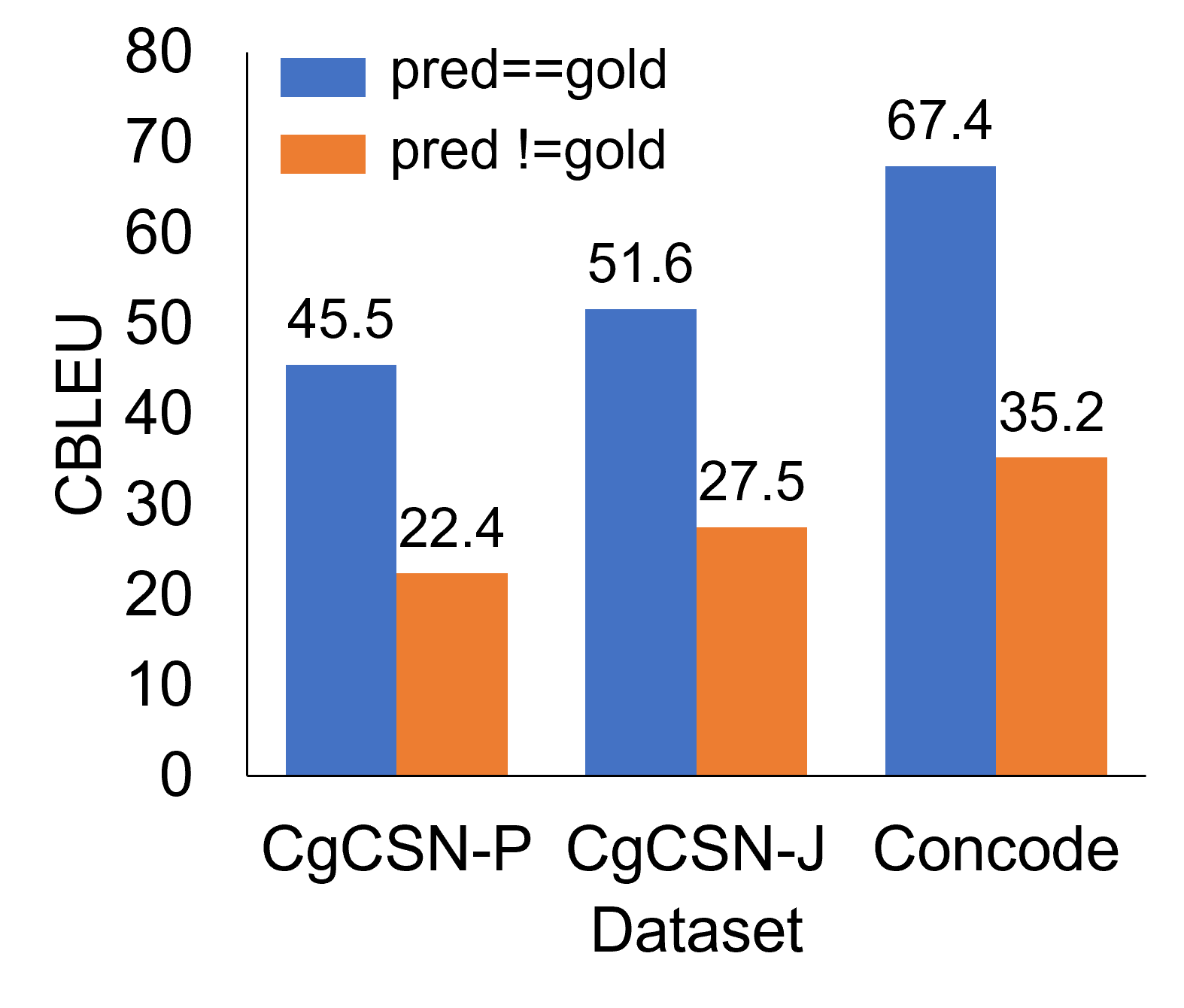}}
    \label{fig:analysis_gen}
    \vspace{0.3mm}
    \subfigure[Code Summarization.] {\includegraphics[width=0.4\linewidth]{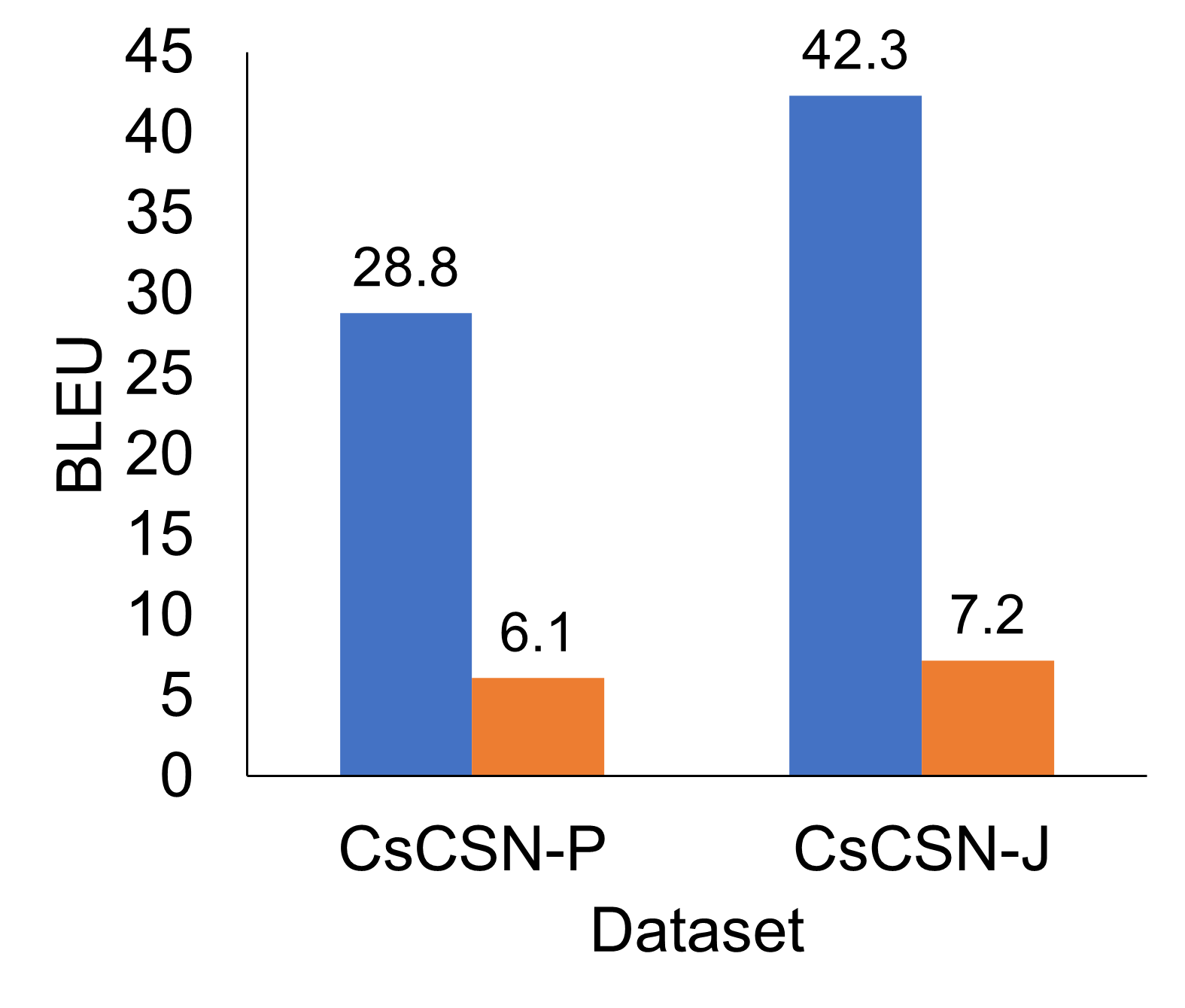}}
    \label{fig:analysis_sum}

    \caption{The Similarity between Top-1 Ranked Code Documents and the Target Answers. Based on whether the model's output matches the target answer, the instances in the testing dataset are divided into two groups (pred$==$gold and pred$!=$gold). Then the CBLEU score between the top-1 ranked code document and the target answer is calculated for each group. The higher CBLEU/BLEU score indicates the top-1 ranked code document is more similar to the target answer, which illustrates the retrieved code document is of high quality to assist the code generation or summarization tasks.}
    \label{fig:analysis_gen_sum}
\end{figure}
\subsection{The Impact of Retrieved Code Snippets on Code-Related Generation Tasks}
In our experiments, we further explore the effectiveness of retrieved code documents in helping CONAN generate code snippets and documentation.

As shown in Figure~\ref{fig:analysis_gen_sum}, we group the datasets of the code generation/summarization tasks into two groups according to whether the prediction result is equal to the golden answer. And we denote the two groups as \texttt{pred$==$gold} and \texttt{pred$!=$gold}. Then we calculate the average CBLEU or BLEU score between the top-1 ranked retrieved code documents and the target answer to estimate the overlap between the retrieved code documents and the golden answers.

Overall, CONAN achieves double CBLEU/BLEU scores when it correctly predicts the golden answers (\texttt{pred$==$gold}), showing that more answer-like code documents can provide the necessary knowledge and guide CONAN-G to generate more accurate codes and summarizations. CONAN-G can refer to and copy some code segments from these retrieved code documents to facilitate the generation process, which supports the motivation of building code retrieval-augmented models in code-related tasks. 
Besides, the \texttt{pred$!=$gold} groups usually achieve higher CBLEU/BLEU scores than the \texttt{pred$==$gold} groups. It demonstrates that these retrieved documents in the \texttt{pred$!=$gold} groups do not help the generation model, which illustrates that the quality of retrieved code documents plays a critical role in guaranteeing the effectiveness of retrieval-augment models.

\subsection{Retrieval Effectiveness of Code Structure Aware Pretraining}
In this experiment, we further evaluate the effectiveness of our code structure aware pretraining method in building a dense retrieval, which includes Code-Documentation Alignment (CDA) and Masked Entity Prediction (MEP). We show the retrieval performance on the code retrieval tasks, then conduct ablation studies, and finally visualize the embedding space.

\textbf{Retrieval Effectiveness in the Code Retrieval Tasks.} We show the effectiveness of our code structure aware pretraining method by evaluating the pretrained models in the code retrieval tasks. In this experiment, we follow previous work~\cite{li2023structure} and use Adv and Codesearch datasets for training and evaluation.

As shown in Table~\ref{tab:retriever}, we start from the code structure aware pretrained model and then finetune the model using different datasets. In the zero-shot setting, CONAN-R outperforms CodeRetriever with about 2\% improvements and 14\% on CodeSearch and Adv, showing the effectiveness of our code structure aware pretraining method. Notably, CONAN-R also shows strong zero-shot ability by achieving comparable performance with the finetuned CodeBERT, GraphCodeBERT, and CodeT5 models. After finetuning, CONAN-R achieves 3.7\% and 9.3\% improvements over CodeT5 on CodeSearch and Adv, respectively. The improvements demonstrate that our pretraining strategy has the ability to enable PLMs to better represent code data and bring its advantages to the downstream code-related retrieval tasks.

\begin{table*}[t]
\centering
\small
\caption{Code Retrieval Performance of CONAN-R. Because of the GPU memory limitation, we set the batch size as 128 during pretraining and finetuning, which is different from previous work~\cite{li2022coderetriever}. All models are evaluated on the CodeSearch and Adv datasets and we report the MRR score.}
\label{tab:retriever}
\begin{tabular}{l |c c c c c c c |c}
\hline
\multirow{2}{*}{\textbf{Model}} & \multicolumn{7}{c|}{\textbf{CodeSearch}} & \multirow{2}{*}{\textbf{Adv}} \\  \cline{2-8} & \textbf{Ruby} & \textbf{Javascript} & \textbf{Go}& \textbf{Python}& \textbf{Java}& \textbf{PHP}& \textbf{Overall}\\ 
\hline
\multicolumn{2}{l}{\textit{\textbf{Zero-Shot}}} \\
\hline
GraphCodeBERT & 1.5 & 0.4 & 0.2 & 0.4 & 0.7 & 2.1 & 0.9 & 0.5\\
CodeRetriever & 68.7 & \textbf{63.7} & 87.6 & 67.7 & \textbf{69.0} & 62.8 & 69.1 & 34.7\\
CONAN-R & \textbf{72.6} & 62.4 & \textbf{88.9} & \textbf{70.0} & 68.6 & \textbf{62.8} &  \textbf{70.9} &  \textbf{46.1} \\\hline
\multicolumn{2}{l}{\textit{\textbf{Fine-Tuning}}} \\
\hline
CodeBERT & 67.9 & 62.0 & 88.2 & 67.2 & 67.6 & 62.8 & 69.3  & 27.2\\
GraphCodeBERT & 70.3 & 64.4 & 89.7 & 69.2 & 69.1 & 64.9 & 71.3 & 35.2\\ 
CodeT5 & 71.9 & 65.5 & 88.8 & 69.8 & 68.6 & 64.5 & 71.5 & 39.3\\
CodeRetriever (Inbatch) & \textbf{75.3} & 69.5 & 91.6 & 73.3 & 74.0 & 68.2 & 75.3 & 43.0\\
CodeRetriever (Hard Negative) & 75.1 & \textbf{69.8} & \textbf{92.3} & \textbf{74.0} & \textbf{74.9} & \textbf{69.1} & \textbf{75.9} & 45.1\\

CONAN-R & 74.7 & 68.6 & 91.8 & 73.7 & 73.7 & 68.6 & 75.2 & \textbf{47.3}\\

\hline
\end{tabular}

\end{table*}


\begin{table}[t]
\centering
\small
\caption{The Retrieval Performance of Ablation Models of Our Code Structure Aware Pretraining Method on Adv Dataset. Masked Entity Prediction (MEP) and Code-Documentation
Alignment (CDA) are two tasks for pretraining CONAN-R, which are proposed by our previous work~\cite{li2023structure}.}
\label{tab:retriever_abla}
\begin{tabular}{c|ccc|cc}
\hline

\multirow{2}{*}{\textbf{Model}} & \multicolumn{3}{c|}{CodeT5} & \multicolumn{2}{c}{CONAN-R}  \\
& Vanilla  & w/ MEP & w/ CDA & Span Mask  & Entity Mask  \\ \hline
\textbf{Zero-Shot} &  0.03  & 0.03   &45.01   & 35.88 & \textbf{46.08}   \\
\textbf{Fine-Tuning} & 39.30   & 38.46 & 46.98 & 42.11  & \textbf{47.28} \\ \hline
\end{tabular}
\end{table}

\textbf{Effectiveness of Different Pretraining Strategies.} Then we explore the effectiveness of pretraining strategies in teaching the CodeT5 model to represent the code for retrieval.

As shown in Table~\ref{tab:retriever_abla}, We start from CodeT5 models and continuously train CodeT5 using two proposed training tasks, Masked Entity Prediction (MEP) and Code-Documentation Alignment (CDA) to show their effectiveness. Meanwhile, we compare the MEP method with the random span masking strategy~\cite{raffel2020exploring,wang2021codet5} to evaluate the effectiveness of different mask modeling strategies. The retrieval performance in both zero-shot and finetuning settings is shown.

Compared with the vanilla CodeT5 model, MEP and CDA show distinct performance in code retrieval. As expected, MEP shows almost the same performance as the baseline model. It shows that only mask language modeling usually shows less effectiveness in learning representations for code data, even using different masking strategies. Different from MEP, CDA shows significant improvements in the code retrieval task. Our CDA training method contrastively trains CodeT5 models using the alignment relations between code and natural language, which helps to bridge the modality gap between them, maps code and natural language in one universal embedding space, and learns more effective representations for retrieval. When adding additional task MEP to CodeT5 (w/ CDA), the retrieval performance of CONAN-R is consistently improved. This phenomenon shows that mask language modeling is still effective in teaching CodeT5 to better capture the structure semantics in code and conduct more effective text representations for code by filling up the masked entities.

We also compare different masking strategies that are used during mask language modeling. Our entity masking strategy outperforms the random span masking strategy, showing the crucial role of entities in code data understanding. Besides, CONAN-R pretrained using the MEP task achieves comparable ranking performance with finetuned models, which illustrates that structure aware pretraining can directly benefit downstream tasks, such as code retrieval.

\begin{figure}
  \centering
  \subfigure[CodeT5.]{\includegraphics[width=0.24\linewidth]{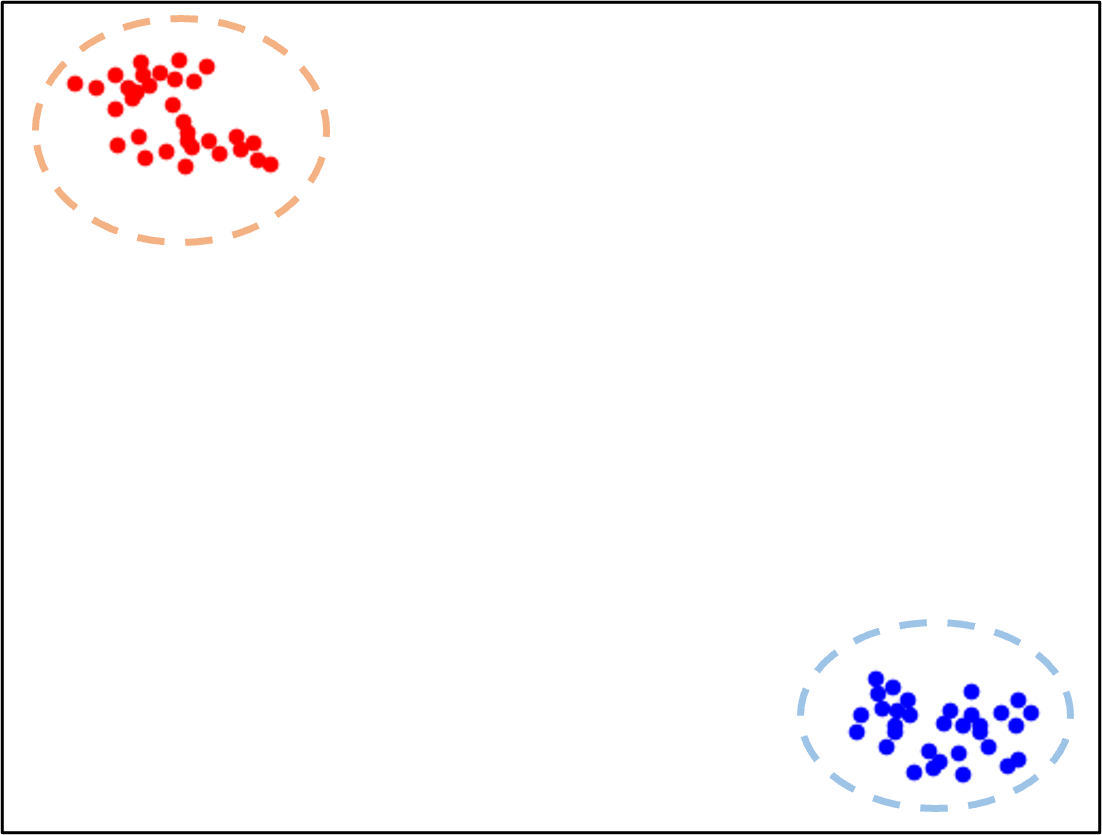}\label{fig:embed:codet5}}
  \subfigure[CodeT5 (w/ CDA).]{\includegraphics[width=0.24\linewidth]{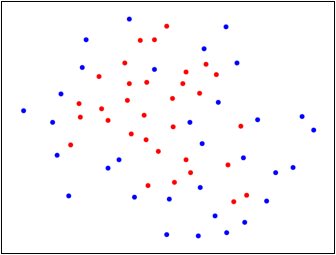}\label{fig:embed:CL}}
  \subfigure[CodeT5 (w/ MEP).]{\includegraphics[width=0.24\linewidth]{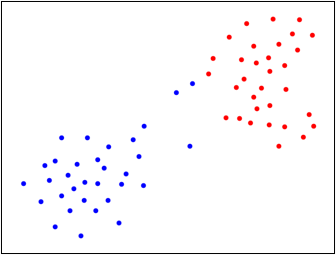}\label{fig:embed:mlm}}
  \subfigure[CONAN.]{\includegraphics[width=0.24\linewidth]{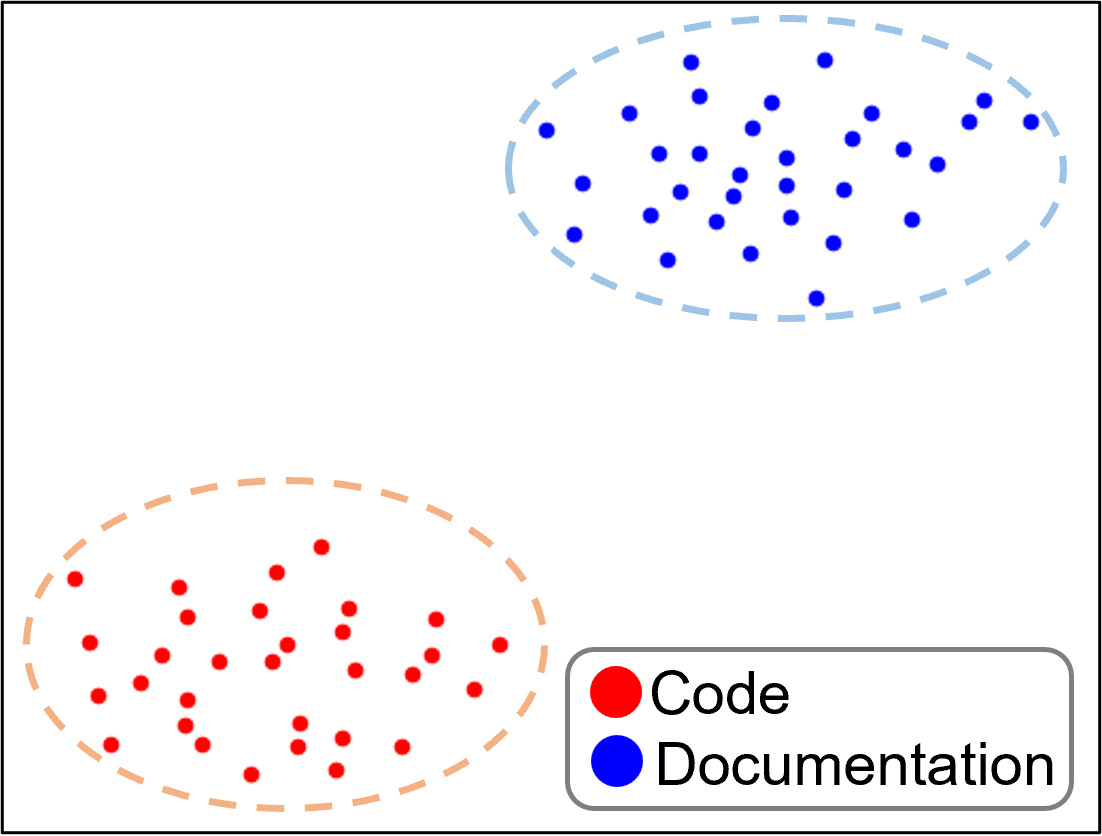}\label{fig:embed:CONAN}}
  
  \caption{Embedding Visualization of Different Models using T-SNE. We randomly sample 32 code snippets (Code) and 32 code documentation (Documentation) from the testing set of the Adv dataset and plot their embedding distribution.}
  \label{fig:embed}
\end{figure}
\begin{table*}[t]
\caption{Case Studies. We randomly sample three cases from code generation, summarization, and completion tasks to show the effectiveness of CONAN. The matched parts are \textcolor[rgb]{0.7,0.3,0.3}{\textbf{emphasized}}. The predictions of CodeT5 are generated directly based on the input, without utilizing any retrieved information.}
\label{tab:case_study}
\small
\scalebox{0.95}{
\begin{tabular}{p{0.95\linewidth}}
\hline

\makecell[c]{\textbf{Code Generation Task}}
\tabularnewline 
\hline
\textbf{Input:} Retrieves connection to Cloud Translate
\tabularnewline
\\

\textbf{Top1-Ranked Code Snippet:} def \_get\_connection ( self ) : if not hasattr ( self , ``\_connection'' ) : if self . use\_pyrax : public = not self . use\_snet \# invert self . \_connection = \textcolor[rgb]{0.7,0.3,0.3}{\textbf{pyrax . connect\_to\_cloudfiles ( public = public )}} elif swiftclient : self . \_connection = swiftclient . Connection ( authurl = self . auth\_url , user = self . username , key = self . api\_key , snet = self . use\_snet , auth\_version = self . auth\_version , tenant\_name = self . auth\_tenant\_name , ) else : raise NotImplementedError ( ``Cloud connection is not correctly configured.'' ) return self . \_connection
\\
\textbf{Top2-Ranked Code Snippet:} def cloudwatch\_connection ( self ) : if self . \_cloudwatc\_connection is None : conn = \textcolor[rgb]{0.7,0.3,0.3}{\textbf{self . \_session . create\_client ( ``cloudwatch'' , self . connection . region )}} self. \_cloudwatch\_connection = conn return self .\_cloudwatch\_connection
\\
\textbf{CodeT5:} def get\_translator ( self , max\_segments = 128 , from\_lang = None ) : if not googleapiclient : raise ImportError ( ``You must install google-apiclient to use translate.'' ) if not self . developer\_key : raise Exception ( ``You must install google-apiclient to use translate.'' ) if not self . service : raise Exception ( ``You must install google-apiclient to use translate.'' ) url\_base = `https://api.googleapis.com/language/translate/v2' params = { `apikey' : self . developer\_key } if from\_lang : params [ `from' ] = from\_lang response = requests . get ( url\_base , params = params ) self . translated\_strings = response . json ( ) return self . translated\_strings
\\
\textbf{CONAN:} def get\_cloud\_translate\_connection ( self ) : if not self . \_cloud\_translate\_connection : self . \_cloud\_translate\_connection = \textcolor[rgb]{0.7,0.3,0.3}{\textbf{CloudTranslateConnectionClient ( credentials = self . \_get\_credentials ( ) )}} return self . \_cloud\_translate\_connection
\\
\textbf{Ground Truth:} def get\_conn ( self ) : if not self . \_client : self . \_client = \textcolor[rgb]{0.7,0.3,0.3}{\textbf{Client ( credentials = self . \_get\_credentials ( ) )}} return self . \_client
\tabularnewline 
\hline

\makecell[c]{\textbf{Code Summarization Task}}
\tabularnewline 
\hline
\textbf{Input:} def  \textcolor[rgb]{0.7,0.3,0.3}{\textbf{create\_instance}} ( self , body , project\_id = None ) : response = \textcolor[rgb]{0.7,0.3,0.3}{\textbf{self . get\_conn ( ) . instances ( )}} . insert ( project = project\_id , body = body ) . execute ( num\_retries = self . num\_retries ) operation\_name = response [ ``nam'' ] self . \_wait\_for\_operation\_to\_complete ( project\_id = project\_id , operation\_name = operation\_name )

\\
\textbf{Top1-Ranked Document:} \textcolor[rgb]{0.7,0.3,0.3}{\textbf{Create an instance}}  within a project.
\\
\textbf{Top2-Ranked Document:} InsertInstance \textcolor[rgb]{0.7,0.3,0.3}{\textbf{creates a new instance}}  on GCP.
\\
\textbf{CodeT5:} \textcolor[rgb]{0.7,0.3,0.3}{\textbf{Create an instance}} within a project.
\\
\textbf{CONAN:} \textcolor[rgb]{0.7,0.3,0.3}{\textbf{Create a new SQL instance.}} 
\\
\textbf{Ground Truth:} \textcolor[rgb]{0.7,0.3,0.3}{\textbf{Creates a new Cloud SQL instance.}} 
\tabularnewline 
\hline

\makecell[c]{\textbf{Code Completion Task}}
\tabularnewline 
\hline
\textbf{Input:} import unittest from contextlib import contextmanager import logging import os from path import path
import shovel import sys class TestRun(unittest.TestCase): ``<STR\_LIT>'' def logs(self, pth, *args, **kwargs): with path(pth): with logs() as out: shovel.run(*args, **kwargs) return [line.strip() for line in out.getvalue().strip().split(`<STR\_LIT:>')] def test\_verbose(self): ``<STR\_LIT>'' actual = self.logs(`<STR\_LIT>', `<STR\_LIT:bar>', `<STR\_LIT>') actual

\\
\textbf{Top1-Ranked Document:} \textcolor[rgb]{0.7,0.3,0.3}{\textbf{Replace}} the current path with the given unencoded path.
\\
\textbf{Top2-Ranked Document:} \textcolor[rgb]{0.7,0.3,0.3}{\textbf{Replace}} absolute urls with relative path.
\\
\textbf{CodeT5:} = [
\\
\textbf{CONAN:} = [line.strip().\textcolor[rgb]{0.7,0.3,0.3}{\textbf{replace}}(os.getcwd(), `<STR\_LIT>') for line in actual]
\\
\textbf{Ground Truth:} = [line.\textcolor[rgb]{0.7,0.3,0.3}{\textbf{replace}}(os.getcwd(), `<STR\_LIT>') for line in actual]
\tabularnewline 
\hline
\end{tabular}
}
\end{table*}

\textbf{Embedding Visualization.} Finally, we present the embedding distribution of documentation texts and their corresponding codes in Figure~\ref{fig:embed}.

Overall, depending on our code structure aware pretraining methods, CONAN conducts a more uniform embedding space than CodeT5 and makes the representations of code snippets and documentation more distinguished in the embedding space. Then we analyze the effectiveness of our continuous training methods, Masked Entity Prediction (MEP), and Code-Documentation Alignment (CDA). By comparing Figure~\ref{fig:embed:CL} with Figure~\ref{fig:embed:codet5}, our code-documentation alignment task indeed helps PLMs to align the representations of code snippets and documentation, which reduces the distance between matched code-documentation pairs and mixes the multi-modal embeddings thoroughly in the embedding space. After adding the masked entity prediction training task to CodeT5 (w/ CDA) (from Figure~\ref{fig:embed:CL} to Figure~\ref{fig:embed:CONAN}), the embedding distributions of code snippets and documentation become distinguished again, demonstrating that masked entity prediction can help models capture different semantics from different data modalities to represent them. Besides, by comparing Figure~\ref{fig:embed:CONAN} with Figure~\ref{fig:embed:mlm}, the code-documentation alignment task also makes the boundary of the embedding clusters of code snippets and documentation clearer. The main reason lies in that these embeddings are assigned to appropriate positions for aligning matched code-documentation pairs with the help of our code-documentation alignment task.

\subsection{Case Study}
Finally, we show three cases from code generation, summarization, and completion tasks to show the effectiveness of CONAN in Table~\ref{tab:case_study}.

For the first case, CONAN retrieves some related code snippets that include some related API/function usages, such as ``connect\_to\_cloudfiles'' and ``create\_clien'', which aims to retrieve a connection to Cloud Translate. These API/function usage examples help the generation CONAN directly implement the ``get\_cloud\_translate\_connection'' function instead of generating some redundant judgment statements. The second case is an example of the code summarization task. In this case, CodeT5 generates a more general summarization ``Create an instance within a project''. CONAN retrieves a general documentation description and a more detailed one, which helps CONAN better understand the summarizations and codes and then generate a specific function summary ``create a SQL instance''. For the code completion case, CodeT5 only generates ``['', showing that the CodeT5 model is not skilled in the code completion task. and CONAN demonstrates its utility in completing the unfinished code. CONAN retrieves some related code documents that replace the path or URL and then correctly generates the golden answer, showing that the related code documents indeed help CONAN complete the unfinished codes. 
\section{Conclusion}
This paper proposes \textbf{CO}de assista\textbf{N}t vi\textbf{A} retrieval-augme\textbf{N}ted language model (CONAN), which aims to help human and LLMs to solve different code-related tasks. CONAN constructs a retrieval-augmented architecture that generalizes to multiple code generation tasks by designing a code structure-aware retriever (CONAN-G) and a dual-view code representation method for building a generation model (CONAN-G). Our experiments show the code document retrieval augmented method is effective in improving the code/documentation generation ability of CodeT5. The improvements derive from a more effective code retriever (CONAN-R) and a better code understanding of the generation model (CONAN-G) by using the code documentation as the code gist. The experimental results on different code-related tasks show the potential advantages of CONAN in building a real-world code assistant by employing the retrieval-augmented generation framework.

\begin{acks}
This work is supported by the Natural Science Foundation of China under Grant (No.  92267201, No. 62206042 and No. U23B2019), the Joint Funds of Natural Science Foundation of Liaoning Province (No. 2023-MSBA-081), and the Fundamental Research Funds for the Central Universities under Grant (No. N2416012).
\end{acks}

\bibliographystyle{ACM-Reference-Format}
\bibliography{citation}

\end{document}